\documentclass[journal]{vgtc}                     

\onlineid{0}

\vgtccategory{Research}

\vgtcpapertype{please specify}

\title{
Adaptive Score Alignment Learning for Continual Perceptual Quality Assessment of 360-Degree Videos in Virtual Reality
}

\author{%
  \authororcid{Kanglei Zhou}{0000-0002-4660-581X},
  \authororcid{Zikai Hao}{0009-0004-5968-7764},
  \authororcid{Liyuan Wang}{0009-0002-7797-325X}, and 
  \authororcid{Xiaohui Liang}{0000-0001-6351-2538}
}

\authorfooter{
  \item Xiaohui Liang is with the State Key Laboratory of Virtual Reality Technology and Systems, Beihang University, Beijing, China, and also with the Zhongguancun Laboratory, Beijing, China. \\
  E-mail: liang\_xiaohui@buaa.edu.cn (\textit{Corresponding author})
  \item Kanglei Zhou and Zikai Hao are with Beihang University, Beijing, China.
  \item Liyuan Wang is with Tsinghua University, Beijing, China.
}

\abstract{%
Virtual Reality Video Quality Assessment (VR-VQA) aims to evaluate the perceptual quality of 360-degree videos, which is crucial for ensuring a distortion-free user experience. 
Traditional VR-VQA methods trained on static datasets with limited distortion diversity struggle to balance correlation and precision. 
This becomes particularly critical when generalizing to diverse VR content and continually adapting to dynamic and evolving video distribution variations.
To address these challenges, we propose a novel approach for assessing the perceptual quality of VR videos, Adaptive Score Alignment Learning (ASAL). 
ASAL integrates correlation loss with error loss to enhance alignment with human subjective ratings and precision in predicting perceptual quality.
In particular, ASAL can naturally adapt to continually changing distributions through a feature space smoothing process that enhances generalization to unseen content. 
To further improve continual adaptation to dynamic VR environments, we extend ASAL with adaptive memory replay as a novel Continul Learning (CL) framework.
Unlike traditional CL models, ASAL utilizes key frame extraction and feature adaptation to address the unique challenges of non-stationary variations with both the computation and storage restrictions of VR devices. 
We establish a comprehensive benchmark for VR-VQA and its CL counterpart, introducing new data splits and evaluation metrics. Our experiments demonstrate that ASAL outperforms recent strong baseline models, achieving overall correlation gains of up to 4.78\% in the static joint training setting and 12.19\% in the dynamic CL setting on various datasets.
This validates the effectiveness of ASAL in addressing the inherent challenges of VR-VQA.
Our code is available at \href{https://github.com/ZhouKanglei/ASAL_CVQA}{https://github.com/ZhouKanglei/ASAL\_CVQA}.
}

\keywords{%
Panoramic video, 360-degree video, virtual reality, video quality assessment, continual learning
}

\teaser{
  \centering
  \begin{overpic}[width=\linewidth]{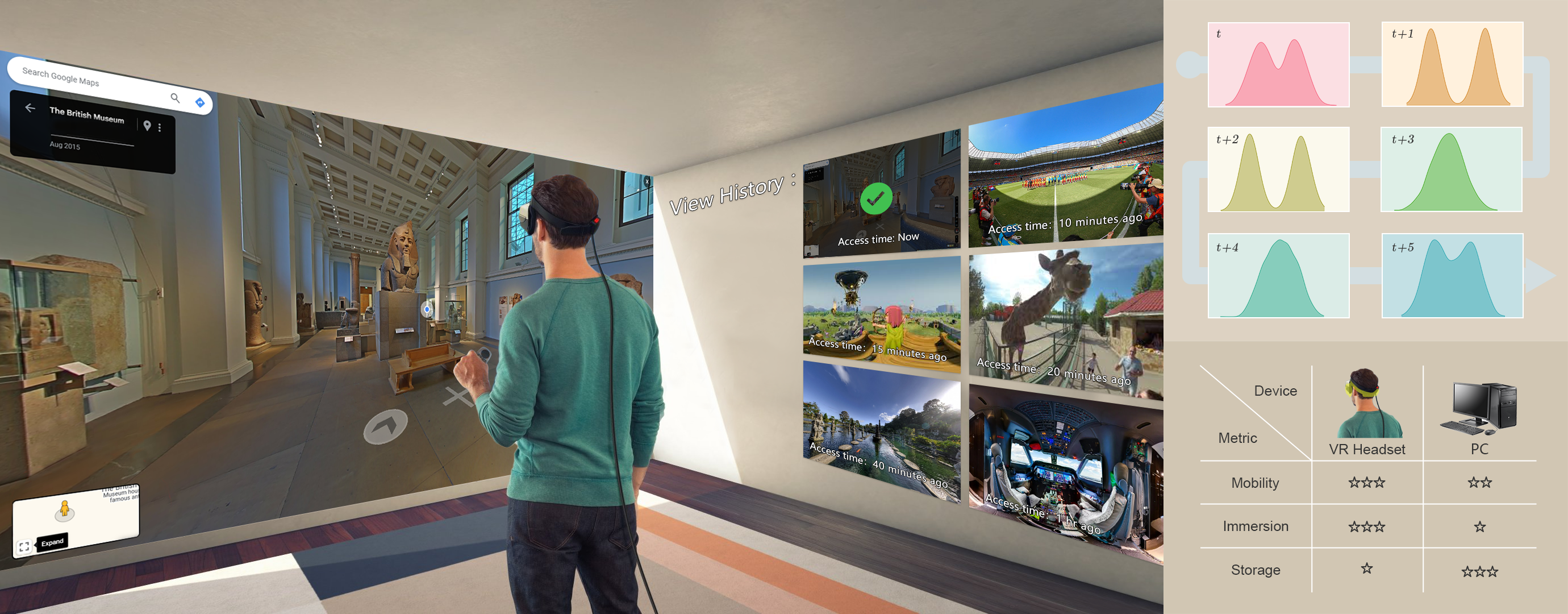}
      \put(1,37.5){\tiny\sf(a)}
      \put(98,37.5){\tiny\sf(b)}
      \put(98,15.5){\tiny\sf(c)}
  \end{overpic}
\caption{
A real-world application scenario:
(a) A user wearing a VR headset is virtually touring the British Museum. User-generated VR videos are often subject to distortions, which can significantly degrade the user experience. Thus, evaluating the perceptual quality of these videos is important.
(b) The video content distribution varies according to individual user preferences. However, traditional models, trained on static datasets, struggle to adapt to these dynamic, non-stationary variations.
(c) The limited computational capacity of VR devices further restricts continuous adaptation for the perceptual assessment of VR videos.
}
\label{fig:teaser}
\phantomsubcaption\label{teaser:a}
\phantomsubcaption\label{teaser:b}
\phantomsubcaption\label{teaser:c}
}

\graphicspath{{figures/}{pictures/}{images/}{./}} 

\usepackage{times}                     

\usepackage{tabu}                      
\usepackage{booktabs}                  
\usepackage{lipsum}                    
\usepackage{mwe}                       
\usepackage{makecell}
\usepackage{pifont}
\usepackage{overpic}
\usepackage{color,colortbl}
\usepackage{amsmath} 
\usepackage{array,multirow,graphicx}

\usepackage{array}
\usepackage{arydshln}
\usepackage{tikz}
\usepackage{pict2e} 

\definecolor{skyblue}{RGB}{0,176,240}
\definecolor{brickred}{RGB}{237,125,49}
\usepackage{contour}

\usepackage{amssymb,bm}
\usepackage[ruled,linesnumbered,vlined]{algorithm2e}

\usepackage[font={sf,scriptsize}]{subcaption,caption}

\makeatletter
\renewcommand\paragraph{\@startsection{paragraph}{4}{1em}%
  {0ex \@plus 1ex \@minus.2ex}%
  {-1em}%
  {\reset@font\normalsize\sffamily\vgtc@sectionfont}}
\makeatother

\begin{document}


\firstsection{Introduction}

\maketitle

360-degree videos, also known as panoramic or immersive videos \cite{anwar2020measuring}, are a significant form of Virtual Reality (VR) streaming. For simplicity, we refer to them as \textit{VR videos} in this paper. These videos offer an immersive experience by allowing users to explore visual scenes in all directions through head-mounted displays. 
This makes them wildly engaged in areas such as education \cite{philippe2020multimodal}, meetings \cite{mcveigh2022beyond}, and medical care \cite{zhou2023video}.
Unlike traditional planar videos, VR videos generally immerse users in a fully immersive virtual environment through headsets (see \cref{teaser:a}), which exacerbates the visibility of minor visual imperfections, such as blurring, stitching errors, or compression artifacts \cite{gao2020quality}. 
These authentic distortions can disrupt the immersive experience \cite{wen2024perceptual} and may even cause physical discomfort \cite{yang20183d}.
Therefore, evaluating the quality of VR videos is crucial to ensure an optimal user experience, especially in applications \cite{zhou2023mixed} where realism and presence are essential.

Traditional VR Video Quality Assessment (VQA) methods \cite{anwar2020measuring,gao2020quality,wen2024perceptual,yang20183d} are adaptations of conventional image and video quality models. However, these methods face significant challenges when applied to complex VR content. 
Specifically, they struggle to achieve an optimal trade-off between precision and correlation while also generalizing effectively across non-stationary content variations.
Firstly, these methods often focus either on correlation with subjective ratings \cite{wen2024perceptual,li2022blindly} or on the precision of predicted quality scores \cite{zhou2023hierarchical,zhou2024magr}. 
In the context of VR, both metrics are indispensable. 
Correlation ensures that the model aligns with user perceptions of overall quality, capturing the subjective experience of immersion. Meanwhile, precision quantifies the severity of specific distortions, which is crucial for identifying and addressing particular quality degradations that disrupt the immersive experience.
Our findings reveal that these objectives are not inherently synchronous. For example, as shown in \cref{fig:loss_lambda}, increasing the weight of the precision loss leads to a decrease in SRCC, despite a significant improvement in precision error. This trade-off indicates that optimizing for one metric can inadvertently compromise the other.
Therefore, models that focus exclusively on either correlation or precision may fail to provide a good user experience in VR. 
Secondly, VR content presents challenges due to its dynamic and evolving nature (see \cref{teaser:b}). Traditional VR-VQA methods are typically trained on static datasets with synthetic distortions and fixed viewing conditions \cite{dataset2018bridge,meng2021viewport}. These static datasets often lack the diversity needed to address the variability in real-world VR content, including differences in user preferences, environmental contexts, and technological factors. 
Consequently, models trained on such datasets struggle to generalize effectively to the continuously changing content distributions in VR environments. 
This limitation complicates the balancing of correlation and precision in VR-VQA, as it reduces the model's ability to maintain global rank consistency (correlation) while accurately predicting individual quality scores (precision).

To address the challenge of balancing precision and correlation in VR-VQA, we propose a novel approach named Adaptive Score Alignment Learning (ASAL). Traditional methods often focus solely on either correlation or precision, potentially overlooking crucial aspects of perceptual quality. 
However, directly integrating these objectives may be suboptimal due to label scarcity and the limited sample size (refer to w/o Repara. in \cref{tab:joint_ablation}). To overcome this, ASAL incorporates a re-parameterization strategy \cite{zhou2023hierarchical} that smooths the feature space, allowing for better generalization across unseen VR content. Then, ASAL uniquely combines correlation loss with error loss, thereby enhancing both the alignment of predicted scores with human subjective ratings and the precision of those predictions. The improved generalization effectively manages the correlation-precision trade-off, providing a robust solution for assessing perceptual quality in VR environments.

Interestingly, we observe that the proposed ASAL can naturally adapt to non-stationary variations in VR-VQA due to improved generalization (refer to the flat minima in \cref{fig:loss_landscape}). 
Moreover, we incorporate Continual Learning (CL) \cite{wang2024comprehensive} to further improve the ability of continual adaptation to real-world variations in VR environments.
CL allows the model to continuously learn from new data without forgetting previously acquired knowledge. This continual adaptation ensures the model maintains an effective balance between precision and correlation, ensuring it remains relevant and effective in real-world VR applications where both content and user interactions are constantly changing.
Unlike other replay-based CL methods \cite{zhou2024magr,li2024continual} that face concerns over storage permission and privacy concerns, our approach in VR can faithfully recover past experiences without these issues. However, optimizing computational and storage efficiency remains essential due to the limited capacity of VR devices (see \cref{teaser:c}).
To this end, we propose adaptive memory replay, which extracts key frames to reduce the storage footprint of replay data and utilizes a feature adapter to reconstruct full video features from these key frames.
The reconstruction in the latent space, rather than the raw data space, ensures computational efficiency.
This adaptation allows ASAL to efficiently manage storage while maintaining the effectiveness of continual quality assessment, thus addressing the challenges of non-stationary content and the computational limitations inherent in VR environments.

As the field of (continual) VR-VQA remains under-explored, we establish a comprehensive benchmark for VR-VQA as well as its CL counterpart. This benchmark includes creating distinct splits of VR-VQA databases, defining custom evaluation protocols and metrics, and incorporating recent strong baselines for comparison. Experiments demonstrate that ASAL significantly outperforms these baseline models, achieving notable average correlation gains of up to 4.78\% in the static joint training (JT) setting and 12,19\% in the dynamic CL setting on various datasets, highlighting its effectiveness in addressing the unique challenges of VR-VQA.
Our contributions are:
\begin{itemize}
 \item We propose a novel adaptive score alignment method that integrates correlation loss and error loss, enhancing the precision of perceptual quality prediction for VR videos.
 \item We validate that the proposed adaptive score alignment can greatly improve the performance of generalization for both JT and CL.
 \item We further develop an adaptive memory replay strategy tailored to CL, utilizing a feature adapter coupled with data compression to effectively adapt the limited storage capacity in VR applications. 
 \item We establish a comprehensive benchmark study for VR video quality assessment by introducing a novel data split strategy and formulating tailored evaluation metrics. 
\end{itemize}

\section{Related Work}
This section first reviews VR video quality assessment. Then, we review continual learning techniques and their applications.


\subsection{VR Video Quality Assessment}
In this study, VR video refers to 360-degree videos in VR for the sake of simplicity.
360-degree videos \cite{pirker2021potential}, also known as panoramic videos, or spherical videos, are recordings that simultaneously capture the entire surrounding environment using an omnidirectional camera or a network of cameras. Existing VR video quality assessment (VQA) methods derive a large part from image and video quality assessment techniques but require adaptation due to the different data formats of VR content.

These methods can be categorized into projected 2D plane, spherical surface, and projected rectilinear viewports.
Planar methods \cite{dataset2018bridge,kim2019deep,sun2017weighted,zakharchenko2016quality} address sphere-to-plane projection issues with techniques such as latitude-dependent weighting \cite{sun2017weighted} and Craster parabolic projection \cite{zakharchenko2016quality}. Spherical methods, such as S-PSNR \cite{yu2015framework} and S-SSIM \cite{chen2018spherical}, aggregate local quality estimates over the sphere. Viewport-based methods, such as those of Xu et al. \cite{xu2020blind} and Li et al. \cite{li2019viewport}, focus on evaluating key viewports, while recent approaches convert panoramic images to planar formats for use with established planar I/VQA methods \cite{sui2021perceptual}. 

Firstly, we observed that some methods in VQA and VR-VQA, such as \cite{wen2024perceptual,li2022blindly}, primarily focus on optimizing the correlation objective, often neglecting precision. 
For instance, Wen et al. \cite{wen2024perceptual} rely solely on a hybrid correlation loss for VR-VQA, achieving high correlation scores but exhibiting significant absolute errors (see \cref{tab:joint_cmp}). 
Such shortcomings highlight the practical limitations of these methods in user-centric applications, where precise quality evaluation is crucial, emphasizing the need for a correlation-precision trade-off.
In contrast, our work differs from related approaches in action quality assessment \cite{zhou2023hierarchical,zhou2024cofinal,zhou2024comprehensivesurveyactionquality}, which predominantly focus on optimizing the precision objective, often overlooking the importance of correlation.

Secondly, these methods often rely on static datasets \cite{wen2024perceptual,meng2021viewport,lopes2018subjective,dataset2018bridge}, which limits their adaptability to dynamic and evolving VR video content. For example, the VOD-VQA dataset \cite{meng2021viewport}, a large panoramic video database, includes only 18 reference videos and 774 distorted videos with limited diversity. This constrained scope hampers the model's ability to generalize across distortions in real-world VR applications. 
Such weak generalization exacerbates the challenge of balancing precision and correlation, highlighting the need for methods that adapt to evolving changes for accurate quality evaluations.

\subsection{Continual Learning}
Continual Learning (CL), also known as lifelong learning, is a field of machine learning focused on enabling models to learn from new data continuously while retaining previously acquired knowledge \cite{wang2021memory,wang2024comprehensive,wang2023incorporating}. This is particularly relevant in scenarios where data evolves, as seen in various applications \cite{graffieti2022continual} including action quality assessment \cite{zhou2024magr,li2024continual,dadashzadeh2024pecop}, semantic segmentation \cite{zhu2023continual,toldo2024learning}, and object detection \cite{liu2023continual,wu2023label}. 

The key challenge of CL is to mitigate catastrophic forgetting \cite{wang2021ordisco,wang2022coscl,wang2023hierarchical}, where a model trained on new tasks rapidly loses its performance on previously learned tasks. This problem arises because traditional machine learning methods typically update the entire model’s parameters based on the most recent data, potentially overwriting information crucial for earlier tasks \cite{van2024continual}.
Representative strategies for mitigating catastrophic forgetting include weight regularization \cite{tang2021gradient}, memory replay \cite{buzzega2020dark,rolnick2019experience,gao2023ddgr}, parameter isolation \cite{james2017ewc}, etc.
These methods aim to balance acquiring new information with retaining existing knowledge, thereby enhancing the model's capacity to learn continuously without degrading its performance on earlier tasks.

Although traditional replay-based methods \cite{buzzega2020dark,rolnick2019experience,zhou2024magr} are effective in mitigating catastrophic forgetting, they present challenges for applications with limited storage capacity, particularly for VR devices. To this end, Wang et al. \cite{wang2021memory} first introduce the data compression concept to reduce the storage cost of representative samples, thereby allowing more representative samples to be stored in the memory buffer. Yang et al. \cite{yang2024probing} propose an effective compression rate selection strategy to balance the quality and quantity of stored samples, mitigating the domain shift issue \cite{janeiro2023visual} that arises when compressed samples degrade classification accuracy.
However, these approaches \cite{wang2021memory,yang2024probing} applied to image data neglect the additional complexity of video data, particularly in VR, where maintaining temporal coherence is essential.
To address these issues, we propose a tailored feature adapter for VR-VQA, designed to reconstruct temporal coherence directly from the latent space.

\section{Definitions of VR-VQA and Continual VR-VQA}
VR-VQA is essential for protecting users from distortion discomfort in immersive environments. However, the scarcity of labeled data results in small datasets with limited distortion diversity, leading to poor generalization. Moreover, models trained on these datasets often fail to adapt to the non-stationary variations encountered in users' everyday lives. To address these challenges, we propose a novel Continual VR-VQA setting incorporating CL strategies in VR-VQA. 
For a clearer comparison, we first define VR-VQA and then introduce its CL counterpart.

\paragraph{Definition of VR-VQA}
VR-VQA aims to predict the perceptual quality score $\hat{s}_i \in \mathbb{R}$ for a given 360-degree video $\mathbf{x} \in \mathbb{R}^{T \times W \times H \times C}$, where $T$, $W$, and $H$ represent the video's temporal length, width, and height, respectively, and $C=3$ corresponds to the RGB color channels. 
The prediction is based on the user's experience in immersive environments using neural networks. The typical neural network architecture used in VR-VQA consists of two main components: a backbone network $f(\cdot)$ for feature extraction and a regressor $g(\cdot)$ for score regression, which together enable the accurate assessment of video quality as perceived by users.
The optimization objective can be expressed as:
\begin{equation} \label{eq:vrvqa_def}
    \min_{\Theta}~~ \mathcal{L}(s_i, \hat{s}_i), ~~ \text{s.t.}~~\hat{s}_i = g(\bm{h}_i), \bm{h}_i = f(\mathbf{x}_i), 
\end{equation}
where $\mathcal{L}$ is the loss function that quantifies the difference between the predicted score $\hat{s}_i$ and the ground truth score $s_i$; it may incorporate scoring error terms \cite{zhou2023hierarchical} or correlation-based constraints \cite{wen2024perceptual}. 
$\mathbf{x}_i$ is the input video, and $\bm{h}_i \in \mathbb{R}^{T\times D}$ represents the hidden feature vector extracted by the backbone network $f(\cdot)$, where $D$ is the dimensionality of the feature space. 
$\Theta = \{\theta_f, \theta_g\}$ represents the learnable parameter set of both the backbone network and the regressor to optimize loss $\mathcal{L}$. 

\paragraph{Definition of Continual VR-VQA}
Continual VR-VQA aims to continuously learn from a series of potentially non-stationary data streams to predict the perceptual quality scores of 360-degree videos. Unlike traditional VR-VQA methods that operate on static datasets, Continual VR-VQA adapts dynamically to evolving datasets $\{\mathcal{D}^t\}_{t=1}^T$, where $T$ represents the total number of learning sessions, with each session referred to as a task.  This approach ensures robust and accurate quality assessment over time. A key challenge is catastrophic forgetting, where learning new data may lead to the loss of previously acquired knowledge. To address this, Continual VR-VQA incorporates a memory bank $\mathcal{M}$ to store and replay a subset of old data, thereby mitigating forgetting while preserving the effectiveness of the model.
Thus, the optimization objective for Continual VR-VQA can be expressed as:
\begin{equation} \label{eq:cvrvqa_def}
    \min_{\Theta}~~\mathcal{L}_{\mathrm{new}} + \alpha \mathcal{L}_{\mathrm{old}},
\end{equation}
where  $\alpha$ balances the trade-off between memory replay on the old knowledge and the adaptation to new data, and $\mathcal{L}_{\mathrm{new}}$ and $\mathcal{L}_{\mathrm{old}}$ denote the loss functions associated with new and old data, respectively. 
Specifically, $\mathcal{L}_{\mathrm{new}}$ measures the model's performance on the current dataset $\mathcal{D}^t$, while $\mathcal{L}_{\mathrm{old}}$ helps prevent forgetting of previously learned information stored in the memory bank $\mathcal{M}$.

This distinction between \cref{eq:vrvqa_def,eq:cvrvqa_def} highlights the additional complexity in Continual VR-VQA, as it necessitates balancing performance on both new and old data. This approach addresses the challenge of catastrophic forgetting and ensures that the model adapts effectively to evolving video quality distributions.

\section{Adaptive Score Alignment Learning (ASAL)}
We first motivate the proposal of our ASAL method and then detail its key novel technical components.

\subsection{Motivation and Framework Overview}
Traditional VR-VQA methods often struggle balance correlation and precision due to static datasets with limited distortion diversity. To address this, we propose ASAL, a novel approach for VR-VQA that enhances both the alignment of predicted scores with human subjective ratings and the precision of perceptual quality predictions. ASAL significantly improves generalization, enabling effective adaptation to unseen VR content. Furthermore, to address the unique challenges in VR, we extend ASAL with adaptive memory replay for Continual VR-VQA. This extension incorporates key frame extraction and feature adaptation to manage storage and computational limitations while maintaining high performance in evolving video distributions.

\begin{figure*}
    \centering
    \includegraphics[width=\linewidth,clip,trim=0 75 0 70]{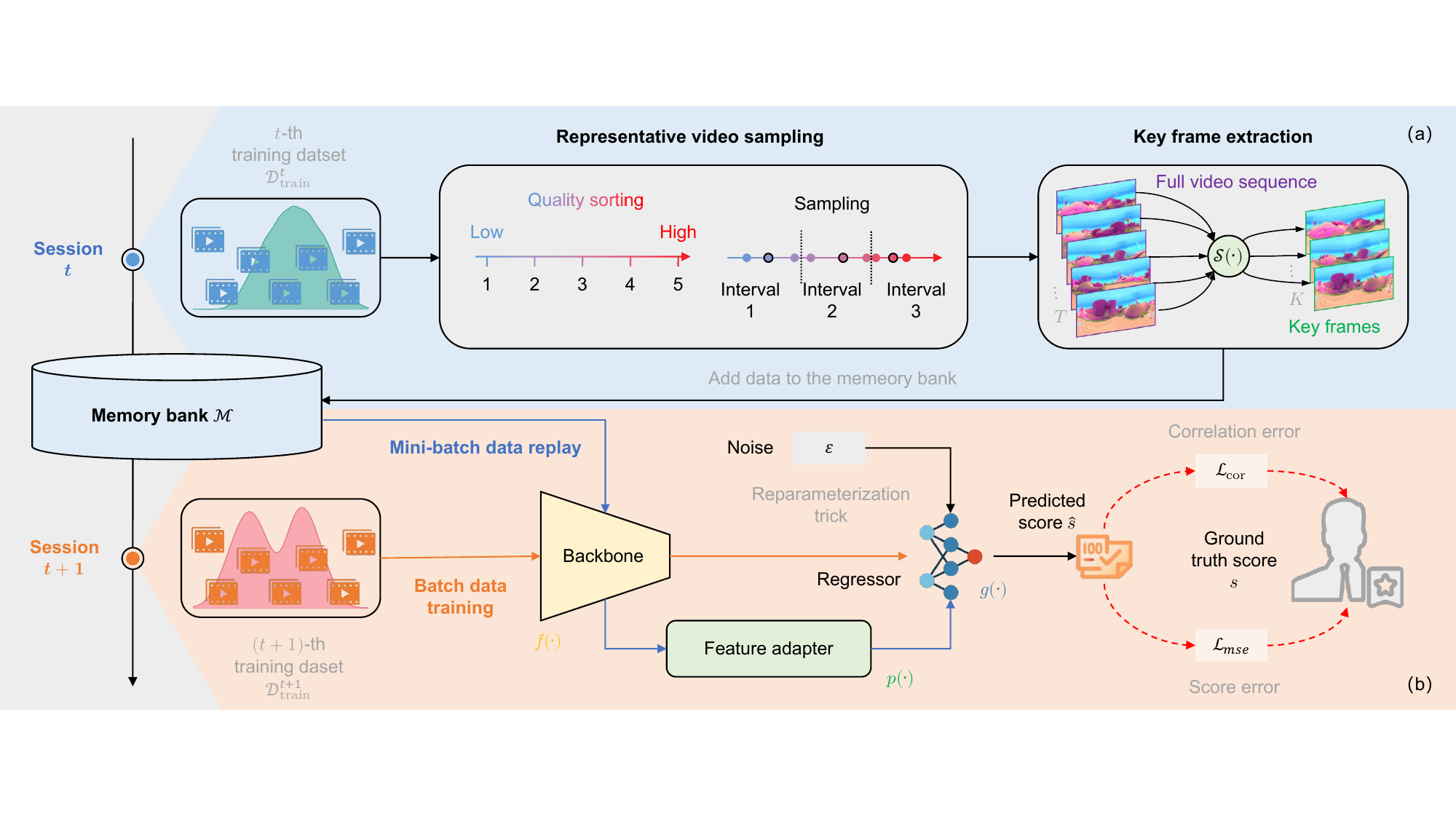}
    \caption{
Overview of our ASAL method:
(a) At the end of task $t$, representative samples are selected and stored in the memory bank. Given the limited storage capacity of VR devices, we further extract key frames from these samples to ensure efficient memory utilization.
(b) During the current task $t$, the current session data and the replay data are processed alternately. For replay, a mini-batch is drawn from the memory bank to extract relevant features. A feature adapter is then used to reconstruct the video sequence before regression. The regressor employs a re-parameterization technique to enhance robustness, and the outputs are optimized using two loss terms to ensure alignment with human assessments.
    }
    \label{fig:framework}
    \phantomsubcaption\label{fig:framework-a}  
    \phantomsubcaption\label{fig:framework-b}
\end{figure*}

\cref{fig:framework} illustrates the whole framework of ASAL, where we consider two consecutive sessions $t-1$ and $t$. In \cref{fig:framework-a}, the process at the end of session $t$ is presented. During this phase, representative samples are selected and stored in a memory bank $\mathcal{M}$. To achieve this, we sort the training dataset from session $t$ based on their quality scores and evenly sample to ensure that the selected examples maintain a distribution similar to the original data.
Given the limited storage capacity of VR devices, we further apply a key frame extraction technique, as suggested by \cite{tang2023deep}, to optimize memory usage. This allows for a greater number of samples to be retained within the same storage constraints, thereby enhancing the model's ability to retain knowledge from previous sessions. 
\cref{fig:framework-b} depicts the process during the current task $t$, where the data of the current session and the replay data are processed alternately. For replay, a mini-batch data is drawn from the memory bank and passed through the backbone network to extract features. 
A feature adapter $p(\cdot)$ reconstructs the full video features from these compressed features.
This reconstruction process enhances computational efficiency by operating in the latent space rather than directly on the raw video data.
The regressor employs a re-parameterization technique to enhance the robustness of the model, and the outputs are optimized using two loss terms $\mathcal{L}_\mathrm{cor}$ and $\mathcal{L}_\mathrm{mse}$ to ensure alignment with human quality assessments.
In this way, our solution to address the optimization problem defined in \cref{eq:cvrvqa_def} can be reformulated as:
\begin{equation} \label{eq:ours_opt}
\begin{aligned}
    \min_{\Theta}~~ &\mathcal{L}_{\mathrm{com}}^{\mathrm{new}} + \alpha \mathcal{L}_{\mathrm{com}}^{\mathrm{old}} + \beta \mathcal{L}_{\mathrm{reg}}, \\
    \text{s.t.}~~&
    \hat{s}_i^t = g(f(\mathbf{x}_i^t)), ~(\mathbf{x}_i^t, s_i^t) \in \mathcal{D}_{\mathrm{train}}^t, \\
    &
    \hat{s}_j = g(p(f(\tilde{\mathbf{x}}_j))), ~(\tilde{\mathbf{x}}_j, s_j) \in \mathcal{M}, \\
\end{aligned}
\end{equation}
where $\mathcal{L}_{\mathrm{com}}^{\mathrm{new}}$ and $\mathcal{L}_{\mathrm{com}}^{\mathrm{old}}$ are combined loss components that balance learning on new data and preserving knowledge from old data (see \cref{eq:l_total}), and $\mathcal{L}_{\mathrm{reg}}$ represents the regularization loss to enhance the feature adaptation process (see \cref{eq:reg}). Here, $\tilde{\mathbf{x}} \in \mathbb{R}^{K\times W\times H \times 3}$ denotes the compressed sample, where $K$ represents the number of compressed frames.  $\beta$ controls the weighting of $\mathcal{L}_{\mathrm{reg}}$.

In the following sections, we first discuss the score alignment learning process in \cref{sec:asl}, which aims to enhance generalization to address the trade-off between correlation and precision. Next, in \cref{sec:amr}, we introduce adaptive memory replay to mitigate catastrophic forgetting with limited computational constraints in CL. 

\subsection{Score Alignment Learning} \label{sec:asl}
Traditional VR-VQA methods \cite{wen2024perceptual,li2022blindly} typically emphasize correlation loss to align predicted scores with human ratings, which may not be sufficient for VR applications. In VR, users require both good alignment with human judgments (correlation) and accurate predictions (precision). While correlation ensures that predictions match human expectations, precision guarantees that these predictions are reliable and accurate. 
However, achieving both high correlation and precision is challenging due to the limited sample size, a consequence of label scarcity.
To address this, we first smooth the feature space to improve the generalization to unseen content, which facilitates a more effective correlation-precision trade-off.
The feature space smoothness step utilizes a re-parameterization trick to improve generalization. This step facilitates a more effective balance between correlation and precision. 

\paragraph{Feature Space Smoothing}
We utilize a probabilistic layer to convert the video-level feature $\bm{h}$ into a random score variable $s$. The variable $s$ is modeled as a Gaussian distribution, defined by:
\begin{equation}
p({s}; \bm{h})=\frac{1}{\sqrt{2 \pi \sigma^2(\bm{h})} } \exp \left(-\frac{(s-\mu(\bm{h}))^{2}}{2 \sigma^{2}(\bm{h})}\right),
\end{equation}
where $\mu(\bm{h})$ represents the mean and $\sigma^2(\bm{h})$ denotes the variance, both dependent on the feature representation $\bm{h}$. Here, the mean $\mu(\bm{h})$ provides the predicted quality score, while the variance $\sigma^2(\bm{h})$ indicates the associated uncertainty.

To efficiently sample from the distribution and integrate this process into training, we employ the re-parameterization trick \cite{kingma2013auto,rezende2014stochastic,kingma2015variational}. This technique generates a random variable $\epsilon$ from a standard normal distribution, $\epsilon \sim \mathcal{N}(0, 1)$, and computes the predicted score $\hat{s}$ as:
\begin{equation}
\hat{s} = \mu(\bm{h}) + \epsilon \cdot \sigma(\bm{h}),
\end{equation}
where $\sigma(\bm{h})$ represents standard deviation. By re-parameterizing the sampling process, we ensure it remains differentiable, which is essential for training the regression network. This technique not only streamlines optimization but also enhances the model’s ability to manage uncertainty, making it more robust in diverse VR-VQA tasks. Readers seeking additional insights into the origins and applications of this method are encouraged to consult \cite{kingma2013auto,rezende2014stochastic,kingma2015variational}.

\paragraph{Correlation-Precision Optimization} 
We introduce a dual-objective optimization strategy that incorporates both correlation and precision losses.
The correlation loss, $\mathcal{L}_{\mathrm{cor}}$, aims to maximize the agreement between the predicted scores and human subjective scores using Pearson's Linear Correlation Coefficient (PLCC), ensuring that the model’s outputs are consistent with human perception, which is:
\begin{equation} \label{eq:plcc}
\mathcal{L}_{\mathrm{cor}} = 1 - \mathrm{PLCC}, ~ \mathrm{PLCC} = \frac{\sum_{i} (s_i - \bar{s})(\hat{s}_i - \bar{\hat{s}})}{\sqrt{\sum_{i} (s_i - \bar{s})^2} \sqrt{\sum_{i} (\hat{s}_i - \bar{\hat{s}})^2}},
\end{equation}
where  $\bar{\hat{s}}$ and ${\bar{s}}$ are the mean values of the predicted and ground truth scores, respectively.
In contrast, the precision loss, $\mathcal{L}_{\mathrm{mse}}$, focuses on minimizing the prediction variance, thereby enhancing the model's reliability and accuracy, which is:
\begin{equation}
    \mathcal{L}_{\mathrm{mse}} = \frac{1}{2N} \sum_{i=1}^N (s_i - \hat{s}_i)^2,
\end{equation}
where $N$ is the total number of training samples.

Our optimization problem can be formulated as a weighted combination of these two objectives:
\begin{equation} \label{eq:l_total}
    \mathcal{L}_{\mathrm{com}} = \mathcal{L}_{\mathrm{cor}} + \lambda \mathcal{L}_{\mathrm{mse}},
\end{equation}
where $\lambda$ is a balancing parameter that adjusts the trade-off between correlation and precision. By adjusting $\lambda$ (refer to the experimental results in \cref{fig:loss_lambda}), we can fine-tune the model to achieve an optimal balance, ensuring that it not only aligns well with human evaluations but also provides accurate and robust predictions. 

This dual-objective optimization ensures a flat loss landscape  (refer to the experimental results in \cref{fig:loss_landscape}). 
A flat loss landscape means that small perturbations in the model's weights lead to minimal changes in the loss function, suggesting that the model has found a robust solution \cite{deng2021flattening}. This robustness often correlates with better generalization, as the model is less sensitive to variations in the data, making it more adaptable to unseen or diverse content.

\paragraph{Benefits} 
These two steps complement each other: the smoother feature space allows easier optimization of the combined loss, leading to more robust and accurate predictions in VR-VQA tasks. Feature space smoothing, using a re-parameterization trick, enhances generalization by stabilizing feature representations, especially with limited data. 
Notably, the enhanced generalization improves performance in both joint training and continual learning. In joint training, it optimizes alignment with human judgments (correlation) and prediction precision. In continual learning, it effectively adapts to non-stationary variations.

\subsection{Adaptive Memory Replay} \label{sec:amr}
Traditional CL methods \cite{zhou2024magr,li2024continual} are often constrained by privacy issues when implementing memory replay. Although privacy concerns are less prominent in VR, the primary limitations arise from the restricted storage space and computational capacity inherent in VR devices, making memory replay more challenging in these settings.
As a result, directly applying established CL methods to VR is not feasible. 
In particular, \cite{wang2021memory} introduced data compression to improve the efficiency of memory replay for images, but is less suitable for videos due to the distinct characteristics and higher complexity of video data. 
Furthermore, video reconstruction introduces additional computational overhead, exacerbating the issue in VR environments. To this end, we propose adaptive memory replay with two essential steps: key frame extraction and feature adaptation. Key frame extraction reduces the storage requirements by selecting representative frames from video data. Subsequently, feature adaptation reconstructs the complete video sequences, allowing for effective learning and adaptation.

\paragraph{Key Frame Extraction}
This process involves selecting a subset of representative frames from a video to capture the essential content while reducing overall memory usage.
Formally, let $\mathcal{X} = \{\bm{x}_1, \bm{x}_2, \ldots, \bm{x}_T\}$ denote a video sequence with $T$ frames. The goal is to select a set of key frames $\tilde{\mathcal{X} } = \{\tilde{\bm{x}}_1, \tilde{\bm{x}}_2, \ldots, \tilde{\bm{x}}_K\}$, where $K < T$, such that these key frames preserve the most critical information for quality assessment.

To achieve this, we define a scoring function $\mathcal{S}(\cdot)$ that evaluates the relevance of each frame based on its contribution to the overall video quality. In practice, this scoring function can be implemented using a pre-trained network (such as ResNet \cite{he2016deep}). This approach leverages the network's ability to capture and quantify critical visual information, providing a robust measure of each frame's relevance. The objective is to maximize the information captured by the selected key frames while minimizing redundancy, which is:
\begin{equation} 
\tilde{\bm{x}}_k = \arg\max_{\bm{x}_i} ~\mathcal{S}(\mathbf{x}_i), \quad \text{for } k = 1, 2, \ldots, K, \end{equation}
where $\tilde{\mathbf{x}}_k$ represents the $k$-th key frame selected from the video sequence. The scoring function $\mathcal{S}(\cdot)$ measures the importance of each frame, often based on factors such as visual content changes or information richness.
To further refine the selection process and ensure diversity among the selected frames, we introduce a diversity constraint that minimizes the similarity between selected frames:
\begin{equation} \min ~\sum_{i \neq j} \mathrm{Sim}(\tilde{\bm{x}}_i, \tilde{\bm{x}}_j), 
\end{equation}
where $\mathrm{Sim}(\cdot, \cdot)$ denotes a similarity measure (e.g., cosine similarity) between pairs of selected key frames. This constraint ensures that the chosen key frames cover different aspects of the video content, improving the effectiveness of the memory replay.

\paragraph{Feature Adaptation}
This step is a crucial component in adaptive memory replay. It aims to reconstruct video sequences from key frames and adapt them to maintain accuracy and relevance in the context of the current task. The process involves two main steps: extracting features from key frames and reconstructing the video sequences.

To achieve this, we use a feature adapter $p(\cdot)$ that operates on the extracted key frames. The feature adapter is designed to map the key frames, which are typically compressed representations, back into a format that is suitable for processing by the regressor network.
This reconstruction process ensures that the video sequences reconstructed from the key frames are as close as possible to the original sequences. The reconstructed sequences are then used as input to the regressor network $g(\cdot)$. Given a compressed sample $\tilde{\mathbf{x}}_j \in \mathbb{R}^{K\times D}$ drawn from the memory bank, we first extract the feature representations $\tilde{\bm{h}}_j$ using the backbone network $f(\cdot)$, which is:
\begin{equation} \label{eq:replay0}
     \tilde{\bm{h}}_j = f(\tilde{\mathbf{x}}_j), \quad \forall \tilde{\mathbf{x}}_j\in \mathcal{M}.
\end{equation}
The feature adapter $p(\cdot)$ then reconstructs the complete video feature $\hat{\bm{h}}_j \in \mathbb{R}^{T\times D}$ from the compressed features $\tilde{\bm{h}}_j$, which is:
\begin{equation} \label{eq:replay1}
    \hat{\bm{h}}_j = p(\tilde{\bm{h}}_j), \quad \forall \tilde{\bm{h}}_j \in \{\tilde{\bm{h}}_j\}_{j=1}^{K}.
\end{equation}
This reconstruction process ensures that the video sequences reconstructed from the key frames are as close as possible to the original sequences. The reconstructed features are then used as input to the regressor network $g(\cdot)$.
To enhance generalization, we design a regularization loss $\mathcal{L}_{\mathrm{reg}}$ by learning by the current data, which is:
\begin{equation} \label{eq:reg}
    \mathcal{L}_{\mathrm{reg}} = \sum_i \| \bm{h}_i^t - \hat{\bm{h}}_i^t \|_2,~~ \text{where} ~\tilde{\bm{h}}_i^t = \phi(\bm{h}_i^t).
\end{equation}
Here, this involves selecting key frames in the feature space, denoted as $\phi(\cdot)$, passing them through the feature adapter, and then reconstructing the full video-level feature $\hat{\bm{h}}_i^t$.
By enforcing this regularization, the future adapter learns to retain critical information while adapting to new tasks, ensuring consistent performance across different data distributions. This step helps maintain the integrity of the learned features and mitigates the risk of overfitting.

\paragraph{Benefits}
Key frame extraction and feature adaptation work collaboratively to enhance efficiency and reduce storage burdens in memory replay. Key frame extraction selects the most representative frames, minimizing the amount of data that needs to be stored and replayed. Feature adaptation further optimizes this process by transforming the selected frames into a more compact and informative feature space, reducing computational overhead. Together, these steps ensure that the most relevant information is retained, allowing for efficient training while alleviating storage and computational constraints in VR.

\section{Benchmark Study}

We benchmark VR-VQA research by contributing a dataset with additional labels, new dataset splits, and new evaluation metrics.

\subsection{Dataset Labelling}
To advance VR-VQA research, we enriched the VRVQW database \cite{wen2024perceptual} by introducing additional category labels.~The VRVQW database is a curated collection of 502 user-generated 360-degree video sequences designed for VR visual quality assessment. The videos vary in frame rates (20-60 fps) and resolutions (1,280 $\times$ 720 to 5,120 $\times$ 2,560 pixels) and are stored in the ERP format after being cropped to 15 seconds. A key feature of VRVQW is its emphasis on authentic distortions common in user-generated VR videos, arising from optical acquisition and stitching challenges. This focus makes the VRVQW database valuable for developing and evaluating VR visual quality models.

The dataset includes six categories of VR scenes, i.e., Cityscape, Landscape, Shows, Sports, CG, and Others. The ``Others'' category covers scenes outside the previous five classes. 
However, the released videos do not originally contain these labels. To address this, each video was manually labeled into the specified categories.  This process resulted in 142 samples for Cityscape, 84 for Landscape, 87 for Shows, 23 for Sports, 34 for CG, and 132 for Others.
Each video has four distinct variants, which are either 7 seconds or 14 seconds in duration, and captured from two viewpoints, labeled A and B. This leads to four combinations: 7A, 7B, 14A, and 14B.

\begin{figure}
    \centering
    \includegraphics[width=\linewidth]{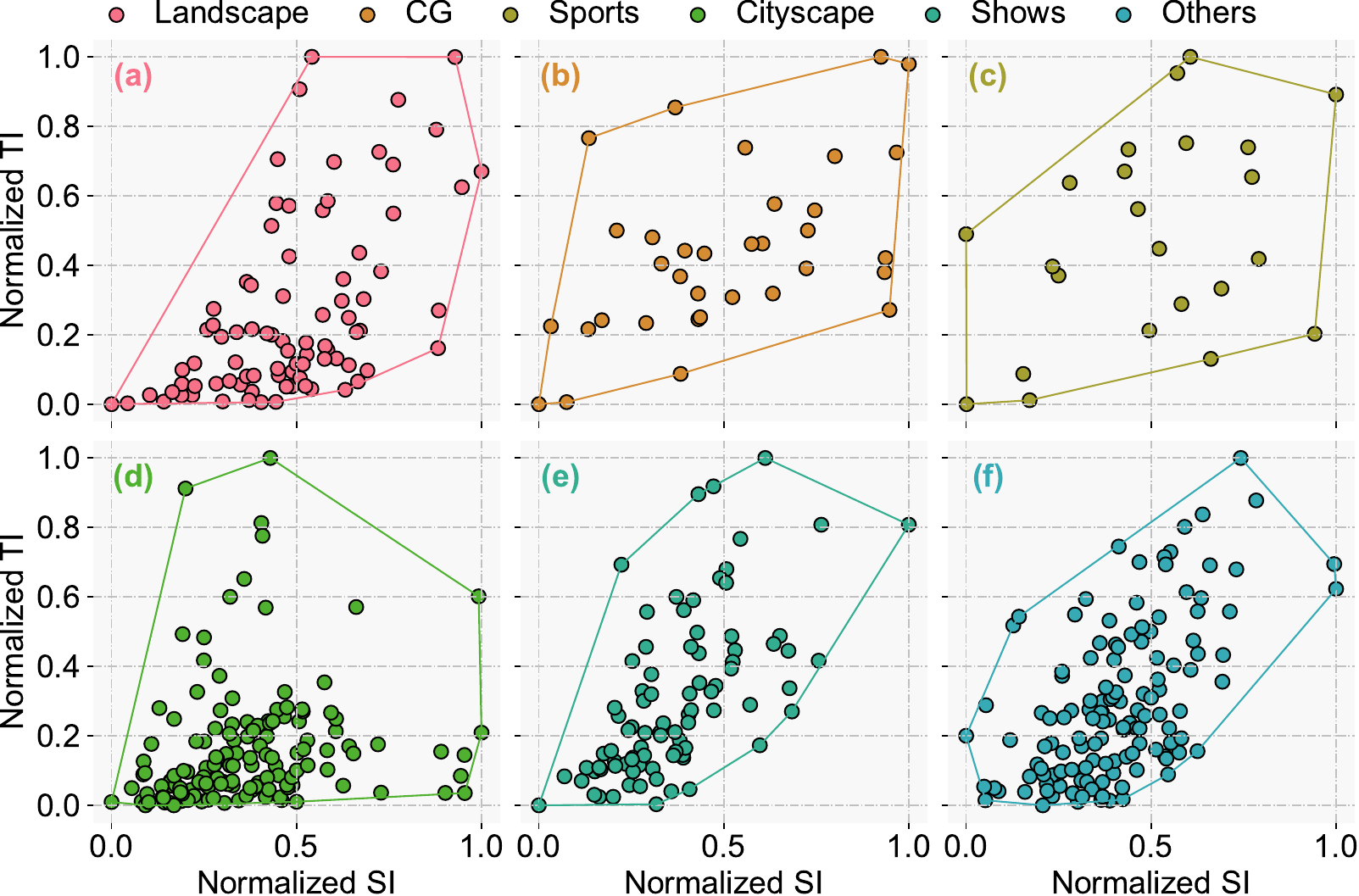}
    \caption{Scatter plots of normalized spatial versus normalized temporal information across different data splits: (a-e) represent sessions 1 to 5, while (f) corresponds to the base session split used for pre-training.}
{
    \phantomsubcaption\label{fig:si_ti_scatter_plots-a}  
    \phantomsubcaption\label{fig:si_ti_scatter_plots-b}
    \phantomsubcaption\label{fig:si_ti_scatter_plots-c}
    \phantomsubcaption\label{fig:si_ti_scatter_plots-d}
    \phantomsubcaption\label{fig:si_ti_scatter_plots-e}
    \phantomsubcaption\label{fig:si_ti_scatter_plots-f}
}%
    \label{fig:si_ti_scatter_plots}
\end{figure}

\subsection{Experimental Protocol}
For the joint training setup, we follow the same experimental protocol as previous methods \cite{wen2024perceptual} to ensure a fair comparison. This consistency across experiments allows for meaningful benchmarking of the proposed approach against existing VR-VQA methods.

For the CL experimental setup, we treat each of the five video categories—Cityscape, Landscape, Shows, Sports, and CG—as a separate task. To show the wide diversity of spatial and temporal richness of the video contents in the database, we calculated the Spatial Information (SI) and Temporal Information (TI) of videos for each session, as shown in \cref{fig:si_ti_scatter_plots}. This task-wise separation ensures that models learn progressively from different types of VR content, facilitating a more comprehensive evaluation of their ability to adapt to new tasks over time. Videos labeled as ``Others'' serve as an auxiliary dataset, used for base session pertaining.  
This allows for a better alignment of domain shift \cite{zhou2024cofinal} without disrupting the integrity of the task-specific learning process.
To address the inherent imbalance in the distribution of video samples across tasks, we apply a uniform sampling strategy, where an equal number of training samples is selected from each task. All samples in the testing sets are maintained, providing a robust evaluation of model performance. This balanced sampling strategy is particularly suited to VR applications, where rapid model updates are essential. By ensuring uniformity in training data, we minimize the risk of any single task dominating the learning process, enabling effective model adaptation across different VR content types.

\subsection{Evaluation Metrics}
Compared to \cite{zhou2024magr}, we propose new CL metrics to comprehensively evaluate the alignment between predicted quality scores and human subjective assessments by capturing both the correlation (ranking) and the precision (magnitude of errors) in model predictions.

For evaluating the static joint training model, we use three key metrics: Pearson's Linear Correlation Coefficient (PLCC), Spearman’s Rank Correlation Coefficient (SRCC), and Relative L2 Error (RL2E), which are used to assess the correlation and precision of quality assessment models. 
Unlike PLCC (see \cref{eq:cvrvqa_def}), which measures the correlation using the actual score values, SRCC provides a better ranking accuracy and is robust to non-linear relationships, which is:
\begin{equation}
    \mathrm{SRCC} = \frac{\sum_i \left(p_i - \bar{p}\right) \left(q_i - \bar{q}\right)}{\sqrt{\sum_{i} \left(p_i - \bar{p}\right)^2 \sum_{i} \left(q_i - \bar{q}\right)^2}},
\end{equation}
where $p_i$ and $q_i$ are the rank orders of the predicted and ground truth scores, respectively, and $\bar{p}$ and $\bar{q}$ are the mean ranks. 
Additionally, RL2E complements SRCC and PLCC by normalizing errors relative to the score range, offering a more precise evaluation of prediction precision across varying scales compared to absolute error metrics.
Given the highest and lowest scores  $s_{\mathrm{max}}$ and $s_{\mathrm{min}}$, $\mathrm{RL2E}$ is: 
\begin{equation}
  \mathrm{RL2E} = \frac{1}{N}\sum_i^N \left( \frac{|s_i - \hat{s}_i|}{s_{\mathrm{max}} - s_{\mathrm{min}}} \right)^2.
\end{equation}

As shown in \cref{tab:joint_cmp}, PLCC and SRCC consistently reflect correlation across all categories. Since both metrics measure correlation, we focus on SRCC for brevity, similar to \cite{zhou2024magr}, which also prioritizes SRCC due to its robustness in evaluating rank-based quality assessments. Additionally, we use RL2E to measure precision, ensuring a comprehensive evaluation of dynamic CL models.
We calculate the overall SRCC, $\mathrm{SRCC}_{\mathrm{ove}}$, by aggregating all samples across sessions rather than averaging the final correlation coefficients for each session. This reduces sensitivity to the number of samples per session. 
The calculation of $\mathrm{RL2E}_{\mathrm{ove}}$ is the same as that of $\mathrm{SRCC}_{\mathrm{ove}}$.

\section{Experiments}
This section introduces implementation details and presents a detailed analysis of experimental results. 

\subsection{Implementation Details}
We implement our ASAL method using PyTorch on a single NVIDIA RTX 3090 GPU with 24 GB of memory. The feature extractor is initialized with a pre-trained model that aligns with the previous work \cite{wen2024perceptual}, while the score regression and adapter networks are randomly initialized.
For optimization, we use the Adam optimizer with an initial learning rate of 0.0001. We set the weight decay parameter to 0.0005 to prevent overfitting.
Each training task is performed for 15 epochs, with a batch size of $b_1 = 3$ and the mini-batch size $b_2 = 2$ for both training and testing. We set the trade-off loss weights $\alpha,\beta=1$ that follows \cite{zhou2024magr}. For the correlation-precision trade-off loss weight $\lambda$ and the buffer size, we conduct comprehensive experiments to refine their optimal values.
The test ratio in each session is 0.2, where only a maximum of 50 samples of them are selected for training.
All other settings follow the protocol outlined in \cite{wen2024perceptual} for a fair comparison.

\subsection{Results and Analysis}
We first present the results of joint training and continual learning models, respectively. Then, we provide qualitative visualizations.

\subsubsection{Joint Training Model Evaluation}

\begin{table}
    \centering
    \setlength{\tabcolsep}{0.38em}
    \caption{Performance comparison of \textbf{joint training} models. $\star$ denotes reimplementation using the official code. 
The \colorbox{cyan!10}{cyan cell} is the best result. For $\uparrow$, higher is better; for $\downarrow$, lower is better.
    }
    \resizebox{\linewidth}{!}{
    \begin{tabular}{rrccccccccc}
    \toprule[0.35mm]
    \multirow{2}{*}[-0.6mm]{Method{\color{gray!75}$^{~\textbf{Year}}$}} & \multicolumn{7}{c}{$\mathrm{PLCC}$ ($\uparrow$)} \\ \cmidrule{2-8}
      &  7A & 7B & 7A\&B & 15A & 15B & 15A\&B & Overall \\
    \midrule
    NIQE \cite{mittal2012making}{\color{gray!75}$^{~12}$} & 0.3220 & 0.4170 & 0.3480 & 0.4090 & 0.3880 & 0.3930 & 0.3710 \\
    VSFA \cite{li2019quality}{\color{gray!75}$^{~19}$} & 0.7200 & 0.8090 & 0.6940 & 0.8250 & 0.8140 & 0.7920 & 0.7460 \\
    MC360IQA \cite{sun2019mc360iqa}{\color{gray!75}$^{~20}$} & \cellcolor{cyan!10}0.7800 & 0.8420 & 0.6640 & 0.8580 & 0.8810 & 0.5200 & 0.5080 \\
    Li et al. \cite{li2022blindly}{\color{gray!75}$^{~22}$} & 0.7340 & 0.7780 & 0.6530 & 0.7600 & 0.7990 & 0.6760 & 0.6690 \\
    HGCN$^{\star}$ \cite{zhou2023hierarchical}{\color{gray!75}$^{~23}$} & 0.4434 & 0.5052 & 0.4305 & 0.4157 & 0.4822 & 0.3893 & 0.4106 \\
    ERP-VQA \cite{wen2024perceptual}{\color{gray!75}$^{~24}$} & 0.7380 & 0.8570 & 0.7300 & 0.8460 & 0.8740 & 0.7910 & 0.7530 \\
    Wen et al. \cite{wen2024perceptual}{\color{gray!75}$^{~24}$} & 0.7550 & \cellcolor{cyan!10}0.8680 & \cellcolor{cyan!10}0.7850 & 0.8600 & 0.8810 & 0.7940 & 0.7610 \\   
    \hdashline
    ASAL (Ours){\color{white!75}$^{~24}$} & 0.7714 & 0.8406 & 0.7617 & \cellcolor{cyan!10}0.8847 & \cellcolor{cyan!10}0.9034 & \cellcolor{cyan!10}0.8340 & \cellcolor{cyan!10}0.7979 \\
    \midrule
    & \multicolumn{7}{c}{$\mathrm{SRCC}$ ($\uparrow$)} \\ \cmidrule{2-8} 
    NIQE \cite{mittal2012making}{\color{gray!75}$^{~12}$} & 0.3500 & 0.4550 & 0.4610 & 0.4290 & 0.3870 & 0.4280 & 0.4070 \\
    VSFA \cite{li2019quality}{\color{gray!75}$^{~19}$} & 0.6940 & 0.7910 & 0.6840 & 0.8240 & 0.8130  & 0.7880 & 0.7500 \\
    MC360IQA \cite{sun2019mc360iqa}{\color{gray!75}$^{~20}$} & 0.7380 & 0.7710 & 0.6420 & 0.7620 & 0.7980 & 0.6750 & 0.6720 \\
    Li et al. \cite{li2022blindly}{\color{gray!75}$^{~22}$} & 0.7640 & 0.8300 & 0.6420 & 0.8530 & 0.8590 & 0.5050 & 0.4380 \\
    HGCN$^{\star}$ \cite{zhou2023hierarchical}{\color{gray!75}$^{~23}$} & 0.4360 & 0.5160 & 0.4339 & 0.4403 & 0.4695 & 0.4049 & 0.4179 \\
    ERP-VQA \cite{wen2024perceptual}{\color{gray!75}$^{~24}$} & 0.7120 & 0.8490 & 0.7230 & 0.8470 & 0.8650 & 0.7820 & 0.7510 \\
    Wen et al. \cite{wen2024perceptual}{\color{gray!75}$^{~24}$} & 0.7300 & \cellcolor{cyan!10}0.8620 & 0.7370 & 0.8590 & 0.8720 & 0.7840 & 0.7570 \\   
    \hdashline
    ASAL (Ours){\color{white!75}$^{~24}$} & \cellcolor{cyan!10}0.7747 & 0.8213 & \cellcolor{cyan!10}0.7572 & \cellcolor{cyan!10}0.8730 & \cellcolor{cyan!10}0.9102 & \cellcolor{cyan!10}0.8343 & \cellcolor{cyan!10}0.7932 \\
    \midrule
    & \multicolumn{7}{c}{$\mathrm{RL2E}$ ($\downarrow$)} \\ \cmidrule{2-8} 
    HGCN$^{\star}$ \cite{zhou2023hierarchical}{\color{gray!75}$^{~24}$} & 0.0458 & 0.0514 & 0.0385 & 0.0635 & 0.0671 & 0.0488 & 0.0392 \\
    ERP-VQA$^{\star}$ \cite{wen2024perceptual}{\color{gray!75}$^{~24}$} & 0.8398 & 1.2425 & 0.8338 & 1.8072 & 0.8962 & 1.0028 & 0.8380 \\
    Wen et al.$^{\star}$ \cite{wen2024perceptual}{\color{gray!75}$^{~24}$} & 1.1571 & 1.3419 & 0.9919 & 1.9986 & 1.1001 & 1.0028 & 0.9797 \\  \hdashline
    ASAL (Ours){\color{white!75}$^{~24}$} & \cellcolor{cyan!10}0.0273 & \cellcolor{cyan!10}0.0289 & \cellcolor{cyan!10}0.0222 & \cellcolor{cyan!10}0.0358 & \cellcolor{cyan!10}0.0166 & \cellcolor{cyan!10}0.0194 & \cellcolor{cyan!10}0.0193\\
    \bottomrule[0.35mm]
    \end{tabular}
    }
    \label{tab:joint_cmp}
\end{table}

\paragraph{Comparison with the State-of-the-Art}
The comparative performance results of joint training models are summarized in \cref{tab:joint_cmp}.
Note that, for a fair comparison, we exclude the scanpath version of \cite{weinan2017proposal} from our evaluation, as it utilizes ground truth data, which would introduce an unfair advantage.
Our proposed method demonstrates superior performance across multiple categories and overall metrics. Specifically, our approach achieves the highest SRCC values in several categories including 7A, 15A, and overall, with values of 0.8162, 0.9073, and 0.7978, respectively. These results highlight our method's effectiveness in aligning predicted scores with human subjective ratings.
In terms of RL2E, our method also outperforms existing models, with the lowest error rates in all categories, including a notable improvement in Overall RL2E with a value of 0.0210. This indicates that our approach not only aligns predictions more accurately but also minimizes prediction errors more effectively compared to previous methods.
In summary, our approach surpasses the second-best model \cite{wen2024perceptual} by approximately 5.3\% in overall SRCC and achieves a 52.5\% improvement in overall RL2E. The significant enhancement in RL2E is largely attributed to considering the correlation-precision trade-off, which effectively balances prediction and alignment with human ratings.

\begin{table}
    \centering
    \setlength{\tabcolsep}{0.38em}
    \caption{Ablation results on our \textbf{joint training} model. 
    }
    \resizebox{\linewidth}{!}{
    \begin{tabular}{lcccccccccc}
    \toprule[0.35mm]
    \multirow{2}{*}[-0.6mm]{Method} & \multicolumn{7}{c}{$\mathrm{PLCC}$ ($\uparrow$)} \\ \cmidrule{2-8}
      &  7A & 7B & 7A\&B & 15A & 15B & 15A\&B & Overall \\
    \midrule
ASAL (Ours) & 0.7714 & 0.8406 & \cellcolor{cyan!10}0.7617 & 0.8847 & \cellcolor{cyan!10}0.9034 & \cellcolor{cyan!10}0.8340 & \cellcolor{cyan!10}0.7979 \\
\hdashline 
~~ w/o $\mathcal{L}_{\mathrm{mse}}$ & 0.7572 & 0.8449 & 0.7488 & 0.8908 & 0.8770 & 0.8188 & 0.7839 \\
~~ w/o $\mathcal{L}_{\mathrm{cor}}$ & 0.7940 & 0.8330 & 0.7341 & 0.8850 & 0.8718 & 0.8106 & 0.7714 \\
~~ w/o Repara. & \cellcolor{cyan!10}0.7945 & \cellcolor{cyan!10}0.8586 & 0.7546 & \cellcolor{cyan!10}0.8956 & 0.8593 & 0.7991 & 0.7759 \\
    \midrule
    & \multicolumn{7}{c}{$\mathrm{SRCC}$ ($\uparrow$)} \\
    \cmidrule{2-8}
ASAL (Ours) & 0.7747 & 0.8213 & \cellcolor{cyan!10}0.7572 & 0.8730 & \cellcolor{cyan!10}0.9102 & \cellcolor{cyan!10}0.8343 & \cellcolor{cyan!10}0.7932 \\
\hdashline 
~~ w/o $\mathcal{L}_{\mathrm{mse}}$ & 0.7810 & 0.8159 & 0.7522 & 0.8889 & 0.9011 & \cellcolor{cyan!10}0.8343 & 0.7898 \\
~~ w/o $\mathcal{L}_{\mathrm{cor}}$ & \cellcolor{cyan!10}0.8116 & 0.8141 & 0.7368 & 0.8798 & 0.8598 & 0.8122 & 0.7707 \\
~~ w/o Repara. & 0.8011 & \cellcolor{cyan!10}0.8553 & 0.7572 & \cellcolor{cyan!10}0.8914 & 0.8649 & 0.7997 & 0.7714 \\
    \midrule
    & \multicolumn{7}{c}{$\mathrm{RL2E}$ ($\downarrow$)} \\ \cmidrule{2-8} 
ASAL (Ours) & 0.0273 & 0.0289 & 0.0222 & 0.0358 & \cellcolor{cyan!10}0.0166 & 0.0194 & 0.0193 \\
\hdashline 
~~ w/o $\mathcal{L}_{\mathrm{mse}}$ & 1.5329 & 1.7493 & 1.3009 & 2.4236 & 1.7265 & 1.5446 & 1.2995 \\
~~ w/o $\mathcal{L}_{\mathrm{cor}}$ & 0.0323 & \cellcolor{cyan!10}0.0216 & 0.0210 & \cellcolor{cyan!10}0.0196 & 0.0308 & \cellcolor{cyan!10}0.0189 & 0.0185 \\
~~ w/o Repara. & \cellcolor{cyan!10}0.0217 & 0.0270 & \cellcolor{cyan!10}0.0194 & 0.0294 & 0.0244 & 0.0201 & \cellcolor{cyan!10}0.0181 \\
    \bottomrule[0.35mm]
    \end{tabular}
    }
    \label{tab:joint_ablation}
\end{table}

\paragraph{Ablation Study}
\cref{tab:joint_ablation} examines the impact of key components of our ASAL method. 
Removing the MSE loss ($\mathcal{L}_{\mathrm{mse}}$) results in a slight reduction in SRCC (from $0.7978$ to $0.7898$) but causes a substantial increase in RL2E (from $0.0210$ to $1.2995$), highlighting the critical role of $\mathcal{L}_{\mathrm{mse}}$ in minimizing prediction errors. Conversely, excluding the correlation loss ($\mathcal{L}_{\mathrm{cor}}$) leads to a more pronounced drop in SRCC (from $0.7978$ to $0.7458$) and a modest increase in RL2E (from $0.0210$ to $0.0392$), indicating its importance in aligning predictions with human subjective ratings.
These results demonstrate that both loss terms play complementary roles in achieving a balance between precision and correlation, critical for effective VR-VQA performance.
The re-parameterization process, when removed, shows a performance drop, with SRCC decreasing from $0.7978$ to $0.7714$ and RL2E increasing from $0.0210$ to $0.0181$. This suggests that re-parameterization contributes to achieving a robust balance between correlation and precision by smoothing the feature space.
In summary, each key component contributes to the overall performance of ASAL. 

\begin{figure}
    \centering
    \includegraphics[width=\linewidth]{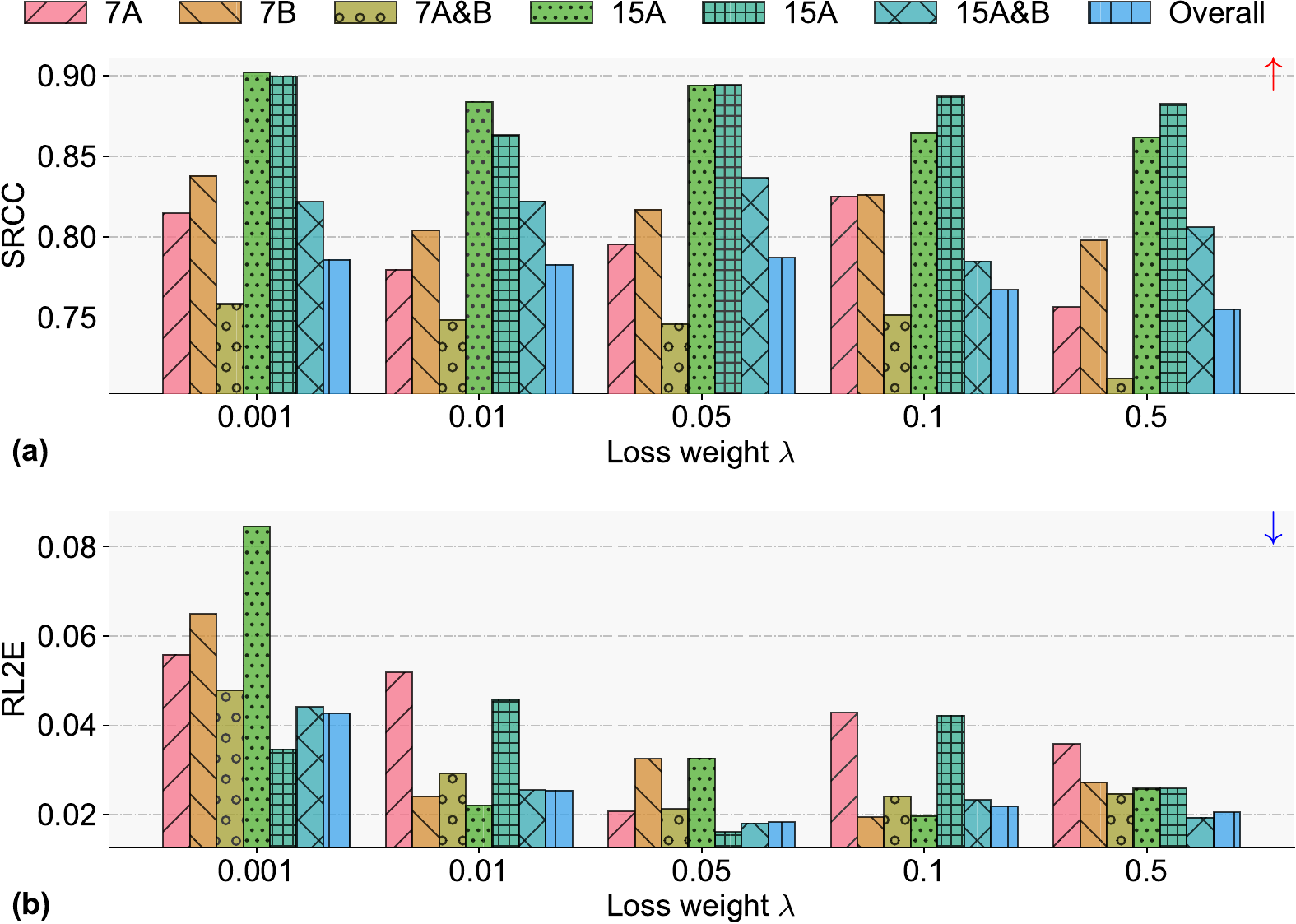}
    \caption{Performance comparison plots of the joint training model with varying loss weights: (a) SRCC ($\uparrow$), (b) RL2E ($\downarrow$).}
    \label{fig:loss_lambda}
\phantomsubcaption\label{fig:loss_lambda-srcc}
\phantomsubcaption\label{fig:loss_lambda-rl2e}
\end{figure}

\paragraph{Impact of Loss Weight $\lambda$}
To explore the impact of the loss weight $\lambda$, \cref{fig:loss_lambda} presents the performance of the joint training model across different values of the loss weight $\lambda$ (see \cref{eq:l_total}), i.e., $0.001$, $0.01$, $0.05$, $0.1$, and $0.5$. 
The results show that increasing $\lambda$ generally results in a reduction of RL2E, indicating an improvement in error minimization, as observed in \cref{fig:loss_lambda-rl2e}.. For example, $\lambda = 0.05$ achieves almost the lowest RL2E values, suggesting it is effective in reducing prediction errors. However, this increase in $\lambda$ induces only slight variations in SRCC, as observed in \cref{fig:loss_lambda-srcc}. This demonstrates that while higher values of $\lambda$ improve error metrics, they do not significantly affect the correlation performance.
Among the tested values, $\lambda = 0.05$ represents a balanced trade-off, providing a good compromise between minimizing RL2E and maintaining SRCC performance. Thus, $\lambda = 0.05$ is identified as the optimal value for achieving an effective balance between correlation and precision in VR-VQA tasks.

\begin{table*}[ht]
    \centering
    \setlength{\tabcolsep}{0.5em}
    \caption{Performance comparison of \textbf{continual learning} models. The \colorbox{cyan!10}{cyan cell} is the best result. For $\uparrow$, higher is better; for $\downarrow$, lower is better.
    }
    \resizebox{\linewidth}{!}{
    \begin{tabular}{lrcccccccccccccccccccccccccc}
    \toprule[0.35mm]
    \multirow{2}{*}[-0.6mm]{Method}  & \multirow{2}{*}[-0.6mm]{Publisher} & \multirow{2}{*}[-0.6mm]{Replay} & \multicolumn{7}{c}{$\mathrm{SRCC}$ ($\uparrow$)} & \multicolumn{7}{c}{$\mathrm{RL2E}$ ($\downarrow$)}   \\ \cmidrule(lr){4-10} \cmidrule(lr){11-17} 
      &  &  & 7A & 7B & 7A\&B & 15A & 15B & 15A\&B & Overall & 7A & 7B & 7A\&B & 15A & 15B & 15A\&B & Overall  \\
    \midrule
    JT & -- & -- & 0.7747 & 0.8213 & 0.7572 & 0.8730 & 0.9102 & 0.8343 & 0.7932 & 0.0273 & 0.0289 & 0.0222 & 0.0358 & 0.0166 & 0.0194 & 0.0193 \\ 
    \midrule
      & &  & \multicolumn{7}{c}{$\mathrm{SRCC}_{\mathrm{ove}}$ ($\uparrow$)} & \multicolumn{7}{c}{$\mathrm{RL2E}_{\mathrm{ove}}$ ($\downarrow$)}   \\ 
      \cmidrule(lr){4-10} \cmidrule(lr){11-17} 
    Seq FT & -- & -- & 0.4400 & 0.3930 & 0.4211 & 0.5233 & 0.4472 & 0.4808 & 0.4567 & 
0.1152 & 0.1247 & 0.0907 & 0.1028 & 0.0712 & 0.0717 & 0.0712 \\
    \hdashline
SI \cite{zenke2017continual} & ICML'17 & \ding{56} &
0.5513 & 0.6928 & 0.6342 & 0.6862 & 0.6114 & 0.6053 & 0.6227 & 
\cellcolor{cyan!10}0.0516 & 0.0473 & 0.0370 & 0.0416 & 0.0444 & 0.0362 & 0.0328 \\
EWC \cite{james2017ewc} & PNAS'17 & \ding{56} &
0.2038 & 0.5030 & 0.3538 & 0.5600 & 0.3160 & 0.4362 & 0.4126 & 
0.1175 & 0.0634 & 0.0654 & 0.0400 & 0.0834 & 0.0534 & 0.0523 \\
LwF \cite{li2017learning} & TPAMI'17 & \ding{56} &
0.4719 & 0.6051 & 0.5393 & 0.6085 & 0.4748 & 0.5373 & 0.5495 & 
0.0625 & 0.0544 & 0.0435 & 0.0400 & 0.0506 & 0.0384 & 0.0364 \\
MER \cite{riemer2019learning} & ICLR'20 & \ding{52} &
0.5029 & 0.4961 & 0.4962 & 0.5133 & 0.5582 & 0.5223 & 0.5190 & 
0.0655 & 0.0646 & 0.0488 & 0.0483 & 0.0545 & 0.0433 & 0.0409 \\
DER++ \cite{buzzega2020dark} & NeurIPS'20 & \ding{52} &
0.4773 & 0.5698 & 0.5309 & 0.7095 & 0.5415 & 0.6155 & 0.5825 & 
0.0652 & 0.0586 & 0.0462 & 0.0323 & 0.0512 & 0.0357 & 0.0359 \\
TOPIC \cite{tao2020few} & CVPR'20 & \ding{52} &
0.4315 & 0.5760 & 0.4897 & 0.6266 & 0.4862 & 0.5283 & 0.5148 & 
0.0593 & 0.0674 & 0.0480 & 0.0573 & 0.0401 & 0.0401 & 0.0389 \\
NC-FSCIL \cite{yang2023neural} & ICLR'23 & \ding{52} &
0.2227 & 0.2632 & 0.2609 & 0.4434 & 0.2704 & 0.3844 & 0.3357 & 
0.0735 & 0.0811 & 0.0585 & 0.0539 & 0.0573 & 0.0467 & 0.0462 \\
SLCA \cite{zhang2023slca} & ICCV'23 & \ding{52} &
0.3983 & 0.3318 & 0.3665 & 0.4534 & 0.2111 & 0.2951 & 0.3290 & 
0.0926 & 0.0765 & 0.0628 & 0.0402 & 0.0896 & 0.0563 & 0.0530 \\
MAGR \cite{zhou2024magr} & ECCV'24 & \ding{52} &
0.4865 & 0.6036 & 0.5386 & 0.6179 & 0.3711 & 0.4852 & 0.5089 & 
0.0707 & 0.0527 & 0.0455 & 0.0371 & 0.0624 & 0.0427 & 0.0394 \\
\hdashline
ASAL (Ours) & -- & \ding{52} & 
\cellcolor{cyan!10}0.6371 & \cellcolor{cyan!10}0.7280 & \cellcolor{cyan!10}0.6936 & \cellcolor{cyan!10}0.7335 & \cellcolor{cyan!10}0.7235 & \cellcolor{cyan!10}0.7149 & \cellcolor{cyan!10}0.7148 & 
0.0517 & \cellcolor{cyan!10}0.0367 & \cellcolor{cyan!10}0.0325 & \cellcolor{cyan!10}0.0308 & \cellcolor{cyan!10}0.0353 & \cellcolor{cyan!10}0.0279 & \cellcolor{cyan!10}0.0267 \\
    \bottomrule[0.35mm]
    \end{tabular}
    }
    \label{tab:cl_cmp}
\end{table*}

\subsubsection{Continual Learning Model Evaluation}
\paragraph{Assessment Comparison with Recent Strong Baselines}
\cref{tab:cl_cmp} presents a comprehensive performance comparison of various CL models. 
Our model outperforms all other methods in both SRCC and RL2E metrics. Specifically, it achieves the highest SRCC scores across all datasets, with an overall SRCC of $0.7148$. This demonstrates superior correlation performance compared to the next best method, which is MAGR \cite{zhou2024magr} with an overall SRCC of $0.5089$. Our model also excels in minimizing error, with the lowest RL2E values across all datasets and an overall RL2E of $0.0267$. This is a significant improvement over the closest competitor, which is SLCA \cite{zhang2023slca} with an overall RL2E of $0.0530$. 
The substantial improvement in RL2E is attributed to our effective consideration of the correlation-precision trade-off, which enhances the model's ability to minimize prediction errors while maintaining high correlation. 
Additionally, the effectiveness of our approach is notably enhanced by two key innovations: key frame extraction and feature adaptation. Key frame extraction helps in efficiently handling non-stationary content by focusing on representative frames, while feature adaptation ensures that our model remains robust and generalizable to diverse and evolving data. These strategies collectively contribute to our superior performance, demonstrating the robust capabilities of our ASAL method in improving perceptual assessment.

\paragraph{Computational Comparison with Recent Strong Baselines}
\cref{tab:computation} presents a detailed comparison of the computational performance of our ASAL method against recent CL baselines. In terms of model parameters, our method introduces a lightweight feature adapter, resulting in an increase of less than 0.01M parameters compared to SLCA \cite{zhang2023slca}, and remains more parameter-efficient than MAGR \cite{zhou2024magr}. Regarding computational complexity, our training time is 0.3000 hours, marginally higher than SLCA (0.2667 hours) but considerably more efficient than MAGR (0.5167 hours). For memory storage, assuming a memory bank size of 50, session count of 5, image resolution of $1024 \times 512$, and a 32-bit float format, our key frame-based approach achieves a storage requirement of 112.50 MB with 3/7 compression, which is substantially lower than DER++ \cite{buzzega2020dark} at 262.50 MB. Notably, while feature-replay-based methods such as SLCA and MAGR minimize storage use, their overall performance in SRCC and RL2E is inferior to ours, underscoring the effectiveness of our approach in balancing computational efficiency and robust continual VR-VQA performance.

\begin{table}[]
    \centering
    \caption{Computational performance comparison.}
    \label{tab:computation}
    \resizebox{\linewidth}{!}{
    \begin{tabular}{lcccccc}
    \toprule
    \multirow{2.5}{*}{Method}
      &  
    \multirow{2.5}{*}{\makecell{Parameter \\ (M)}} & 
    \multirow{2.5}{*}{\makecell{Training \\ Time (h)}} & 
    \multirow{2.5}{*}{\makecell{Storage \\ (MB)}} &
    \multicolumn{2}{c}{Overall Performance} \\ \cmidrule(lr){5-6} 
    & & & &
    $\mathrm{SRCC}_{\mathrm{ove}}$ &
    $\mathrm{RL2E}_{\mathrm{ove}}$ \\ 
    \midrule
    DER++ \cite{buzzega2020dark} & 11.34 & 0.6167  & 262.50 & 0.5825 & 0.0359\\
    SLCA \cite{zhang2023slca}    & 11.34 & 0.2667  & ~~~~0.63 & 0.3290 & 0.0530  \\
    MAGR \cite{zhou2024magr}     & 11.44 & 0.5167  & ~~~~0.01 & 0.5089 & 0.0394\\ 
    \hdashline
    ASAL (Ours) & 11.34 & 0.3000 & 112.50 & 0.7148 & 0.0267 \\
    \bottomrule
    \end{tabular}
    }
\end{table}

\begin{table}[ht]
    \centering
    \caption{Ablation results on our \textbf{continual learning} model. 
    }
    \resizebox{\linewidth}{!}{
    \begin{tabular}{lrcccccccccccccccccccccccccc}
    \toprule[0.35mm]
    \multirow{2.5}{*}[-0.6mm]{Method} & \multicolumn{7}{c}{$\mathrm{SRCC}_{\mathrm{ove}}$ ($\uparrow$)}  \\ \cmidrule(lr){2-8}
& 7A & 7B & 7A\&B & 15A & 15B & 15A\&B & Overall \\
\midrule
ASAL (Ours) &
0.6371 & \cellcolor{cyan!10}0.7280 & \cellcolor{cyan!10}0.6936 & 0.7335 & 0.7235 & 0.7149 & \cellcolor{cyan!10}0.7148  \\
\hdashline
~~w/o Key Frame &
0.6429 & 0.7106 & 0.6779 & 0.7448 & 0.7326 & 0.7157 & 0.7062  \\
~~w/o Feature Adapter &
\cellcolor{cyan!10}0.6487 & 0.6976 & 0.6728 & 0.7295 & 0.6800 & 0.6791 & 0.6846  \\
~~w/o $\mathcal{L}_{\mathrm{reg}}$ &
0.6213 & 0.7045 & 0.6650 & \cellcolor{cyan!10}0.7682 & \cellcolor{cyan!10}0.7358 & \cellcolor{cyan!10}0.7329 & 0.7105 \\
~~w/o Base Pre-Train &
0.3664 & 0.5035 & 0.4525 & 0.6394 & 0.4512 & 0.4877 & 0.4805 \\
\midrule
  &  \multicolumn{7}{c}{$\mathrm{RL2E}_{\mathrm{ove}}$ ($\downarrow$)}   \\ \cmidrule(lr){2-8} 
ASAL (Ours) &
\cellcolor{cyan!10}0.0517 & \cellcolor{cyan!10}0.0367 & \cellcolor{cyan!10}0.0325 & 0.0308 & \cellcolor{cyan!10}0.0353 & \cellcolor{cyan!10}0.0279 & \cellcolor{cyan!10}0.0267 \\
\hdashline
~~w/o Key Frame  &
0.0566 & 0.0409 & 0.0359 & 0.0383 & 0.0381 & 0.0320 & 0.0302 \\
~~w/o Feature Adapter &
0.0626 & 0.0383 & 0.0368 & 0.0268 & 0.0493 & 0.0327 & 0.0309 \\
~~w/o $\mathcal{L}_{\mathrm{reg}}$ &
0.0660 & 0.0370 & 0.0374 & \cellcolor{cyan!10}0.0219 & 0.0471 & 0.0299 & 0.0296 \\
~~w/o Base Pre-Train & 
0.0817 & 0.0585 & 0.0516 & 0.0411 & 0.0613 & 0.0437 & 0.0421 \\
    \bottomrule[0.35mm]
    \end{tabular}
    }
    \label{tab:ablation}
\end{table}
\begin{figure*}[h]
    \centering
    \includegraphics[width=\linewidth]{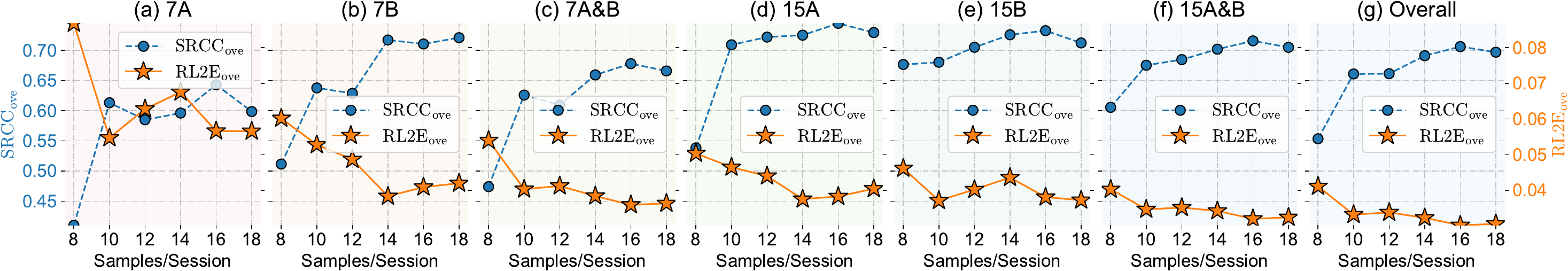}
    \caption{
Line plots for the impact of representative samples per session on the overall performance metrics ($\mathrm{SRCC_{ove}}$ and $\mathrm{RL2E_{ove}}$).
    }
    \label{fig:impact_buffer_size}
\end{figure*}

\paragraph{Ablation Study}
\cref{tab:ablation} shows the effect of key components in our model. The removal of key frame extraction results in a slight decrease in $\mathrm{SRCC_{ove}}$ from 0.7148 to 0.7062, representing a reduction of 1.2\%, and an increase in $\mathrm{RL2E_{ove}}$ from 0.0267 to 0.0302, a 13.1\% degradation in error measurement accuracy. 
Notably, key frame extraction proves to be beneficial for memory stability due to redundancy removal.
When the feature adapter is removed, the $\mathrm{SRCC_{ove}}$ drops more significantly to 0.6846, a reduction of 4.2\%, indicating the effectiveness of feature reconstruction. The exclusion of $\mathcal{L}_{\mathrm{reg}}$ leads to a smaller decrease in $\mathrm{SRCC_{ove}}$ (0.7105, a 0.6\% reduction) but a more notable increase in $\mathrm{RL2E_{ove}}$ to 0.0296 (10.9\% increase), indicating using current data to learn the reconstruction dependencies is beneficial. Finally, removing base pre-training results in the largest performance drop, with $\mathrm{SRCC_{ove}}$ plummeting by 32.8\% to 0.4805, confirming the effectiveness of using the auxiliary dataset for domain gap adaptation \cite{zhou2024cofinal}. These results confirm the critical role of each component for robust perceptual quality assessment.

\begin{figure}
    \centering
    \includegraphics[width=\linewidth]{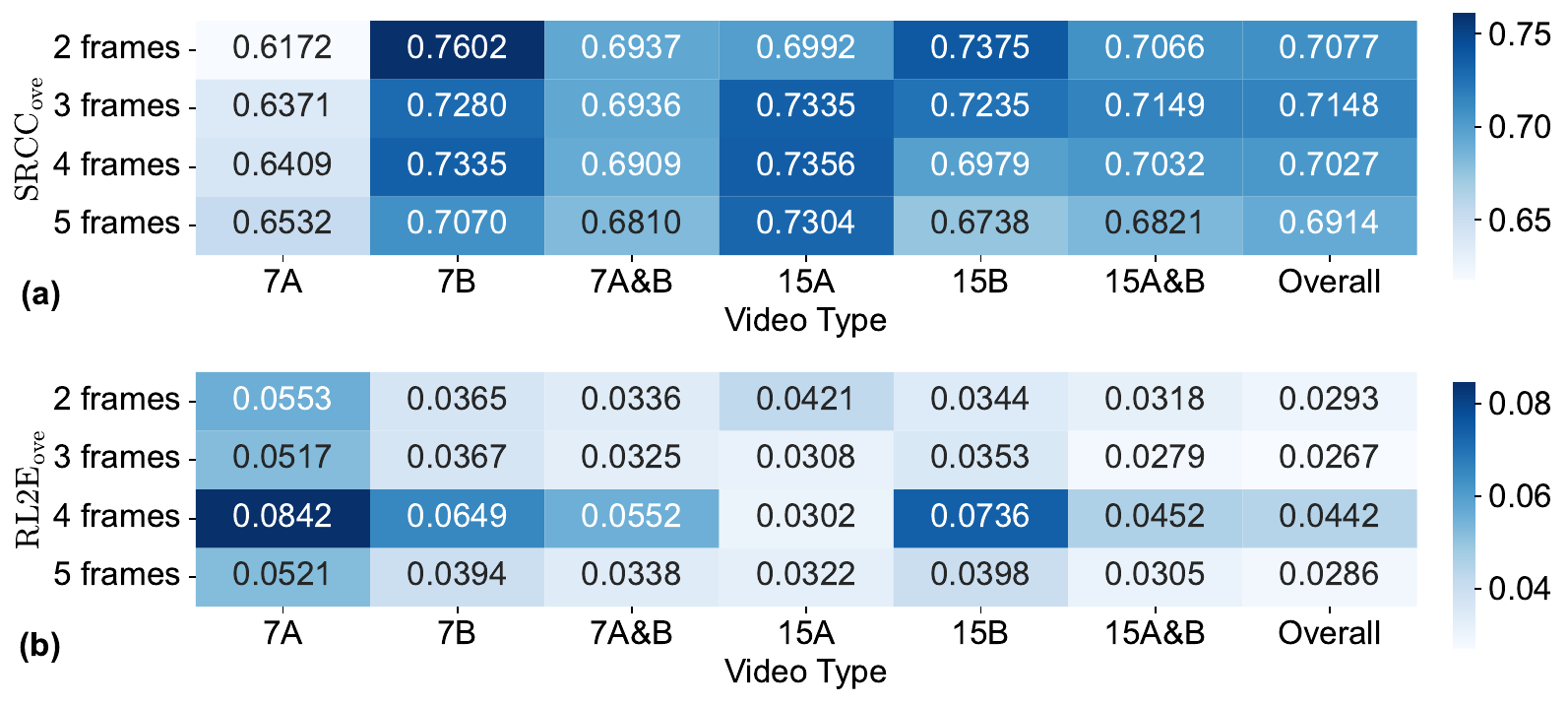}
    \caption{Performance comparison heatmaps of our model with varying key frames: (a) $\mathrm{SRCC_{ove}}$ ($\uparrow$), (b) $\mathrm{RL2E_{ove}}$ ($\downarrow$).}
    \label{fig:compression_frames}
\phantomsubcaption\label{fig:compression_frames-srcc}
\phantomsubcaption\label{fig:compression_frames-rl2e}
\end{figure}

\begin{figure*}
    \centering
    \sf
    \small
    \setlength{\tabcolsep}{0.05em}
    \resizebox{\linewidth}{!}{
    \begin{tabular}{ccccccc}
    \cellcolor{gray!20}
    & 
    \cellcolor{gray!20}The first frame & \cellcolor{gray!20}The middle frame & \cellcolor{gray!20} The last frame & 
    \cellcolor{gray!20} Model output \\
    \cellcolor{gray!20}\rotatebox{90}{~~~\color{skyblue}(a) Sample \#0016}
    & \cellcolor{gray!20}
    \begin{overpic}[width=0.28\linewidth]{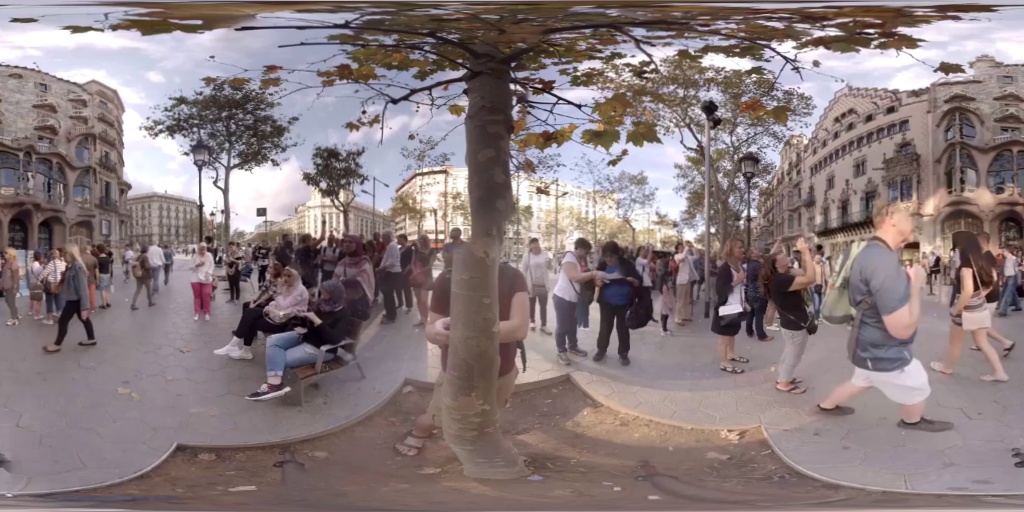}
    \put(40,3){\linethickness{0.25mm}\color{yellow}\polygon(0,0)(15,0)(15,10)(0,10)}
    \put(40,5){\linethickness{0.25mm}\color{yellow}\polygon(0,15)(15,15)(15,25)(0,25)}
    \put(26,42){\linethickness{0.25mm}\color{yellow}\line(1,-2.05){14.1}}
    \put(26,42){\linethickness{0.25mm}\color{yellow}\line(1,-0.85){14.1}}
    \put(5,42){\color{yellow}\contour{black}{Abrupt structure change}}
    \end{overpic}
    & \cellcolor{gray!20}
    \begin{overpic}[width=0.28\linewidth]{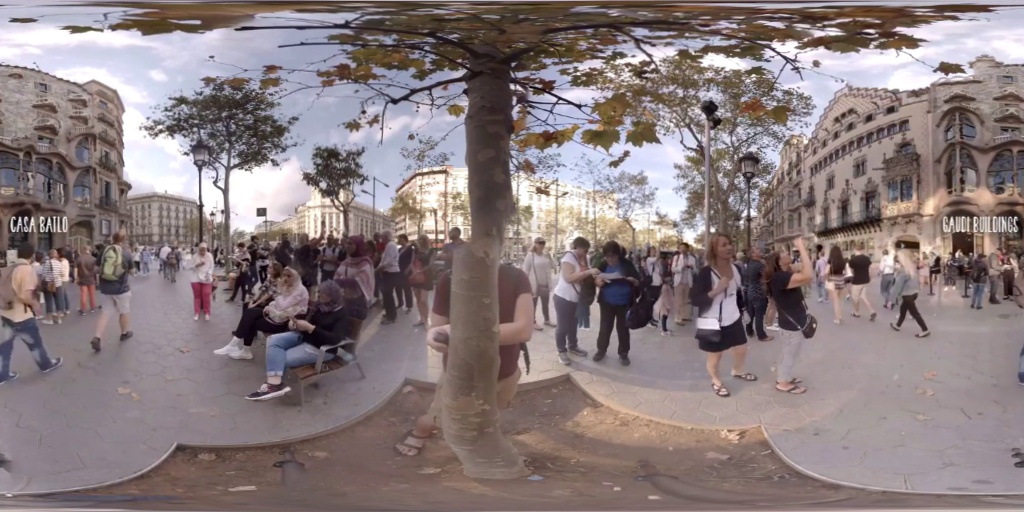}
    \put(40,25){\linethickness{0.25mm}\color{green}\polygon(0,0)(15,0)(15,10)(0,10)}
    \put(80,20){\linethickness{0.25mm}\color{green}\polygon(0,0)(12,0)(12,10)(0,10)}
    \put(26,42){\linethickness{0.25mm}\color{green}\line(1,-0.22){54}}
    \put(26,42){\linethickness{0.25mm}\color{green}\line(1,-0.5){15}}
    \put(5,42){\color{green}\contour{black}{Over-exposure}}
    \end{overpic}
    & \cellcolor{gray!20}
    \begin{overpic}[width=0.28\linewidth]{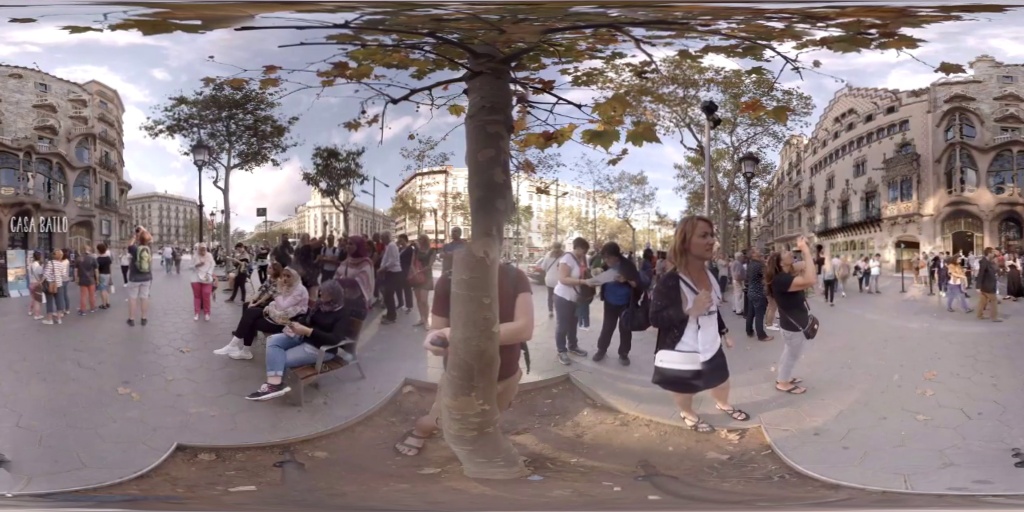}
    \put(60,5){\linethickness{0.25mm}\color{yellow}\polygon(0,0)(15,0)(15,10)(0,10)}
    \put(26,42){\linethickness{0.25mm}\color{yellow}\line(1,-0.8){33.5}}
    \put(5,42){\color{yellow}\contour{black}{Abrupt structure change}}
    \end{overpic}
    & \cellcolor{gray!20}
    \begin{overpic}[width=0.14\linewidth,height=0.14\linewidth]{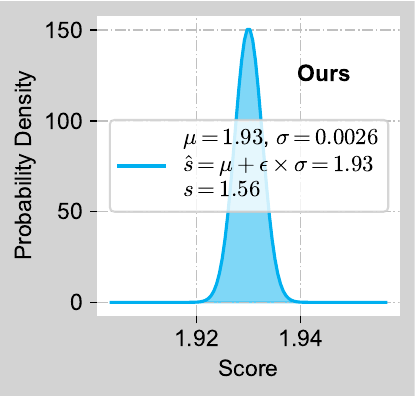}
    \end{overpic}
    \\
    \cellcolor{gray!20}\rotatebox{90}{~~~\color{skyblue}(b) Sample \#1413}
    & \cellcolor{gray!20}
    \begin{overpic}[width=0.28\linewidth]{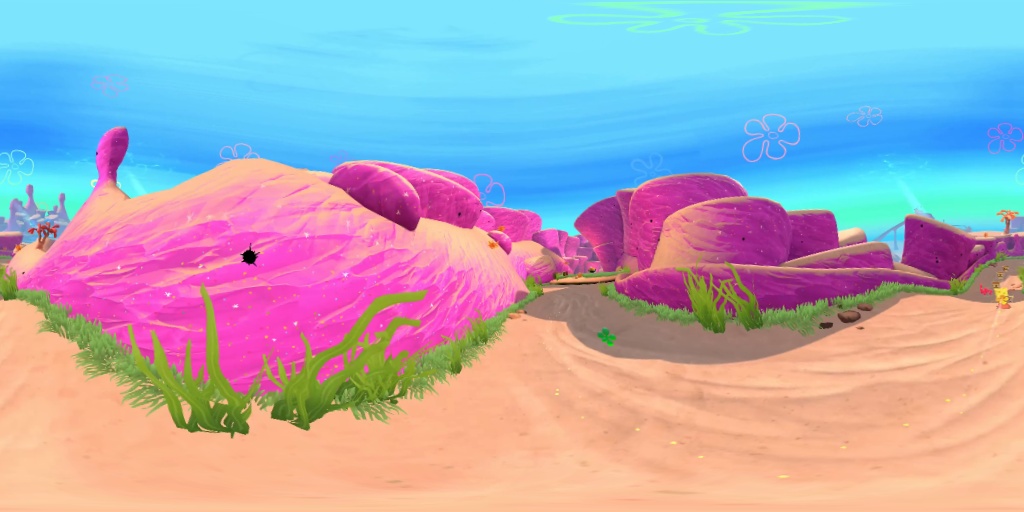}
    \put(5,42){\color{white}\contour{black}{No obvious distortion}}
    \end{overpic}
    & \cellcolor{gray!20}
    \begin{overpic}[width=0.28\linewidth]{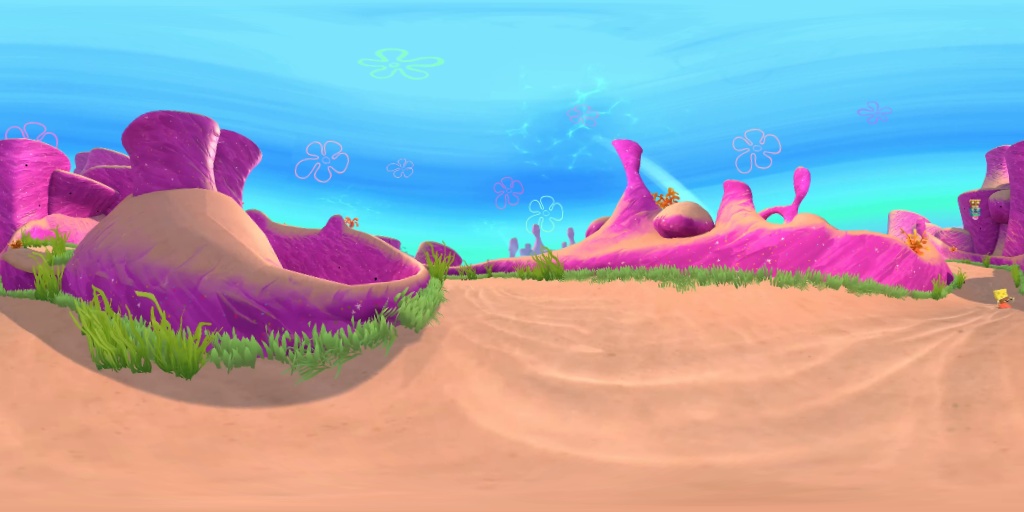}
    \put(5,42){\color{white}\contour{black}{No obvious distortion}}
    \end{overpic}
    & \cellcolor{gray!20}
    \begin{overpic}[width=0.28\linewidth]{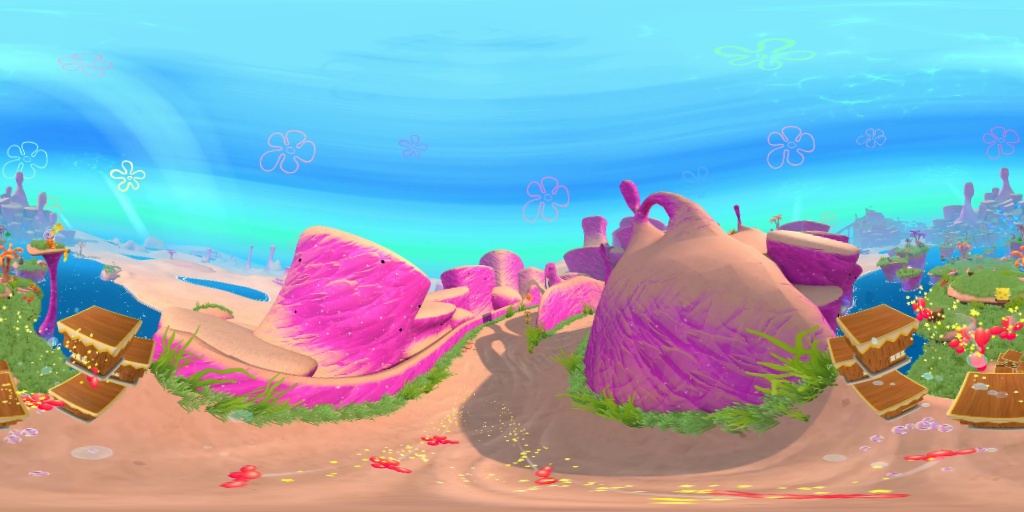}
    \put(5,42){\color{white}\contour{black}{No obvious distortion}}
    \end{overpic}
    & \cellcolor{gray!20}
    \begin{overpic}[width=0.14\linewidth,height=0.14\linewidth]{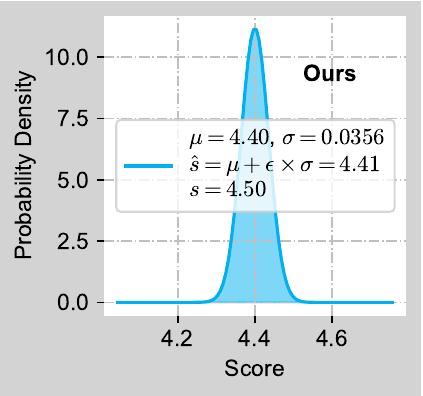}
    \end{overpic}
    \\
    \cellcolor{gray!20}\rotatebox{90}{\color{brickred}~~~(c) Sample \#1800}
    & \cellcolor{gray!20}
    \begin{overpic}[width=0.28\linewidth]{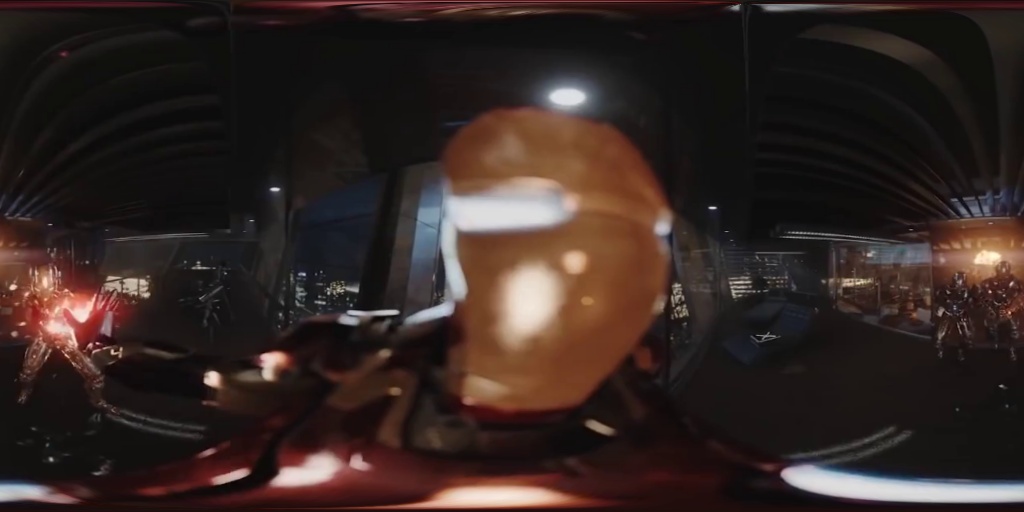}
    \put(40,10){\linethickness{0.25mm}\color{magenta}\polygon(0,0)(10,0)(10,25)(0,25)}
    \put(26,42){\linethickness{0.25mm}\color{magenta}\line(1,-0.46){15}}
    \put(5,42){\color{magenta}\contour{black}{Object with missing parts}}
    \end{overpic}
    & \cellcolor{gray!20}
    \begin{overpic}[width=0.28\linewidth]{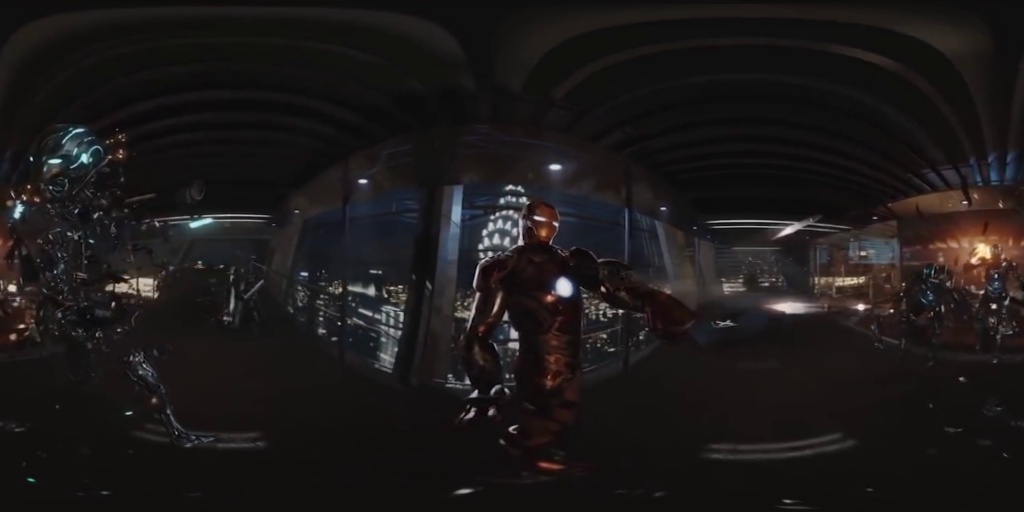}
    \put(0,10){\linethickness{0.25mm}\color{yellow}\polygon(0.5,0)(18,0)(18,30)(0.5,30)}
    \put(65,42){\linethickness{0.25mm}\color{yellow}\line(-5,-2){47}}
    \put(40,42){\color{yellow}\contour{black}{Abrupt structure change}}
    \end{overpic}
    & \cellcolor{gray!20}
    \begin{overpic}[width=0.28\linewidth]{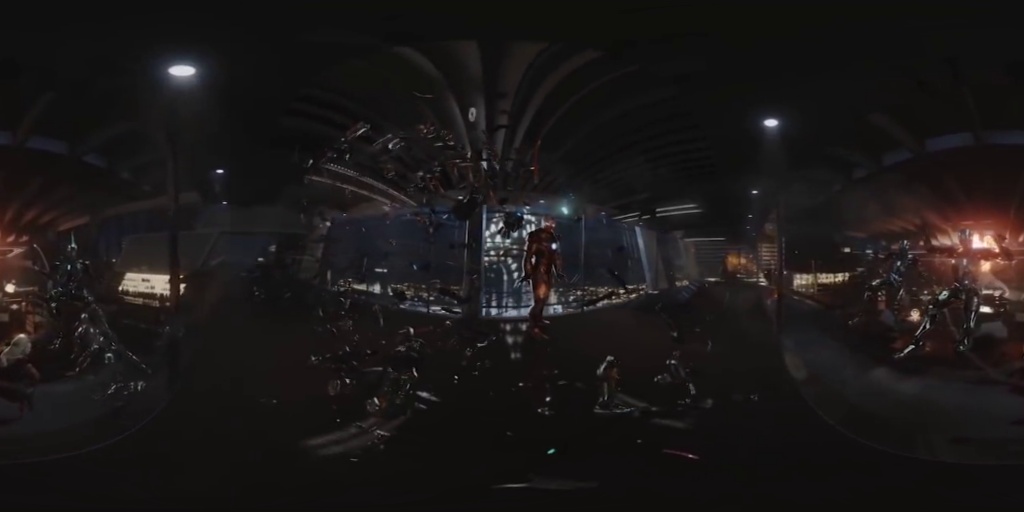}
    \put(5,42){\color{cyan}\contour{black}{Under-exposure}}
    \end{overpic}
    & \cellcolor{gray!20}
    \begin{overpic}[width=0.14\linewidth,height=0.14\linewidth]{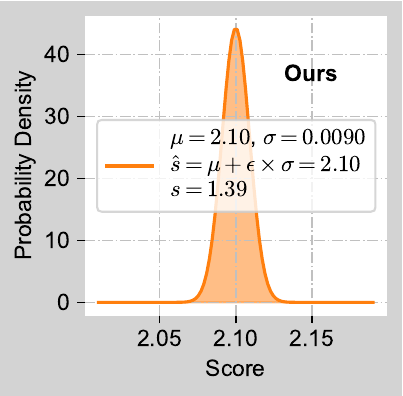}
    \end{overpic}
    \\
    \end{tabular}
    }
    \caption{
Qualitative results: (a) and (b) show better performance of ours, while (c) shows poor performance from both ours and \cite{wen2024perceptual}. 
}
    \label{fig:case_study}
\phantomsubcaption\label{fig:case_study-a}
\phantomsubcaption\label{fig:case_study-b}
\phantomsubcaption\label{fig:case_study-c}
\end{figure*}


\paragraph{Impact of Buffer Size}
The line plots in \cref{fig:impact_buffer_size} illustrate the impact of varying the number of representative samples per session (set to 8, 10, 12, ..., 18) on overall performance.
Initially, as the sample size increases from 8 to 16, both $\mathrm{SRCC_{ove}}$ and $\mathrm{RL2E_{ove}}$ show improvement, indicating that a larger buffer size helps in capturing more representative data, leading to enhanced performance. However, beyond 16 samples, performance begins to decline. This drop suggests that while a larger sample buffer generally benefits the model, excessively large buffers may introduce noise or lead to overfitting, adversely affecting performance. Therefore, a trade-off exists, where optimal performance is achieved with a balanced buffer size, typically around 16 samples, to maximize effectiveness while minimizing potential downsides.

\paragraph{Impact of the Number of Key Frames}
\cref{fig:compression_frames} illustrate the performance of our model under varying key frames (2-5). For $\mathrm{SRCC_{ove}}$ (see \cref{fig:compression_frames-srcc}), higher values are generally observed with moderate compression ratios, indicating improved performance in perceptual quality assessment. In terms of $\mathrm{RL2E_{ove}}$ (see \cref{fig:compression_frames-rl2e}), lower values are achieved with moderate compression ratios, reflecting better accuracy. 
Notably, using 3 key frames achieves the best overall results, optimizing both perceptual quality and error estimation. This suggests that 3 key frames provide an ideal balance between maintaining compression efficiency and maximizing model performance, making it a practical choice for VR-VQA in evolving environments.


\begin{figure*}[!h]
\begin{minipage}{.32\textwidth}
    \centering
    \includegraphics[width=\linewidth]{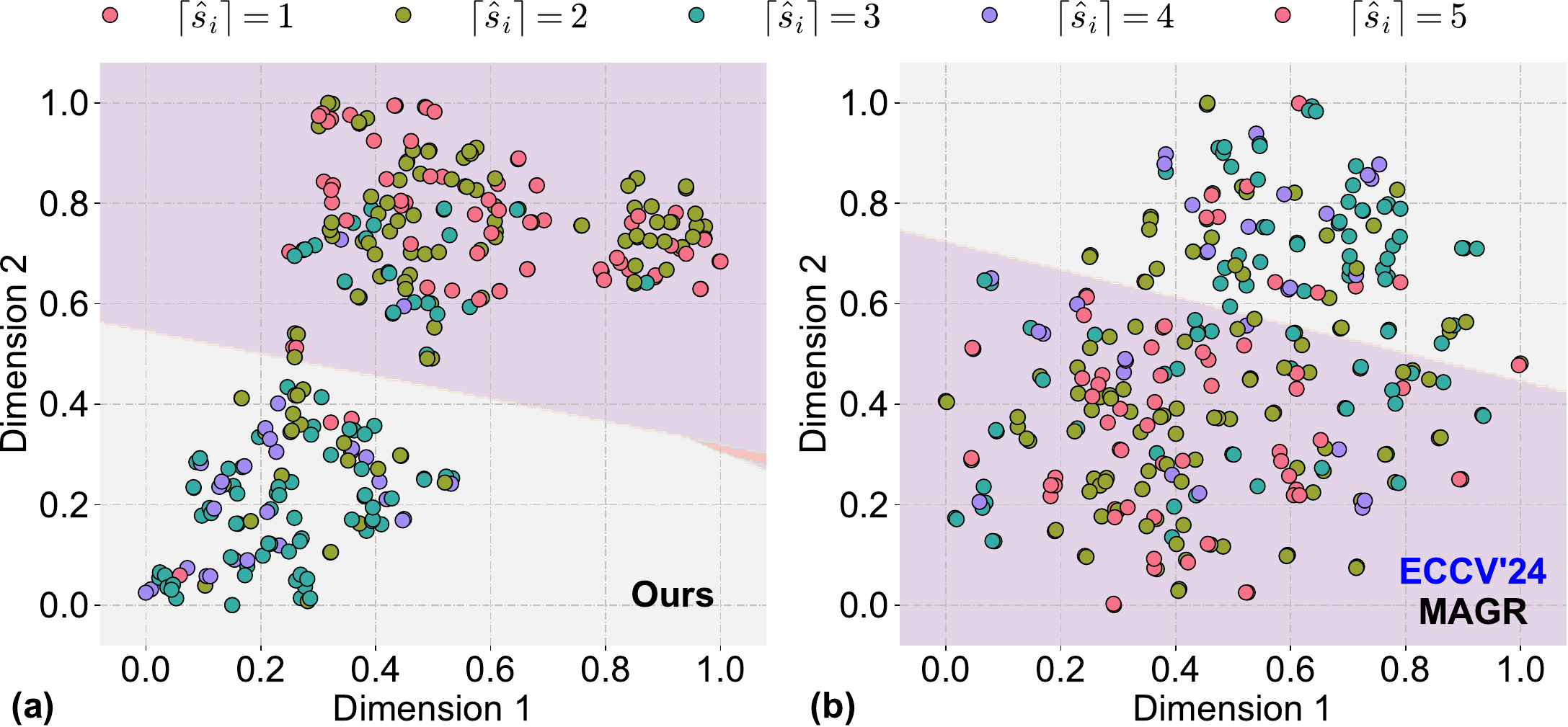}
    \caption{
Feature scatter plots with score discretization generated using the t-SNE toolbox: (a) Ours, (b) MAGR.
    }
    \label{fig:tsne_cmp}
\phantomsubcaption\label{fig:tsne_cmp-a}
\phantomsubcaption\label{fig:tsne_cmp-b}
\end{minipage}~
\begin{minipage}{.32\textwidth}
    \centering
    \includegraphics[width=\linewidth]{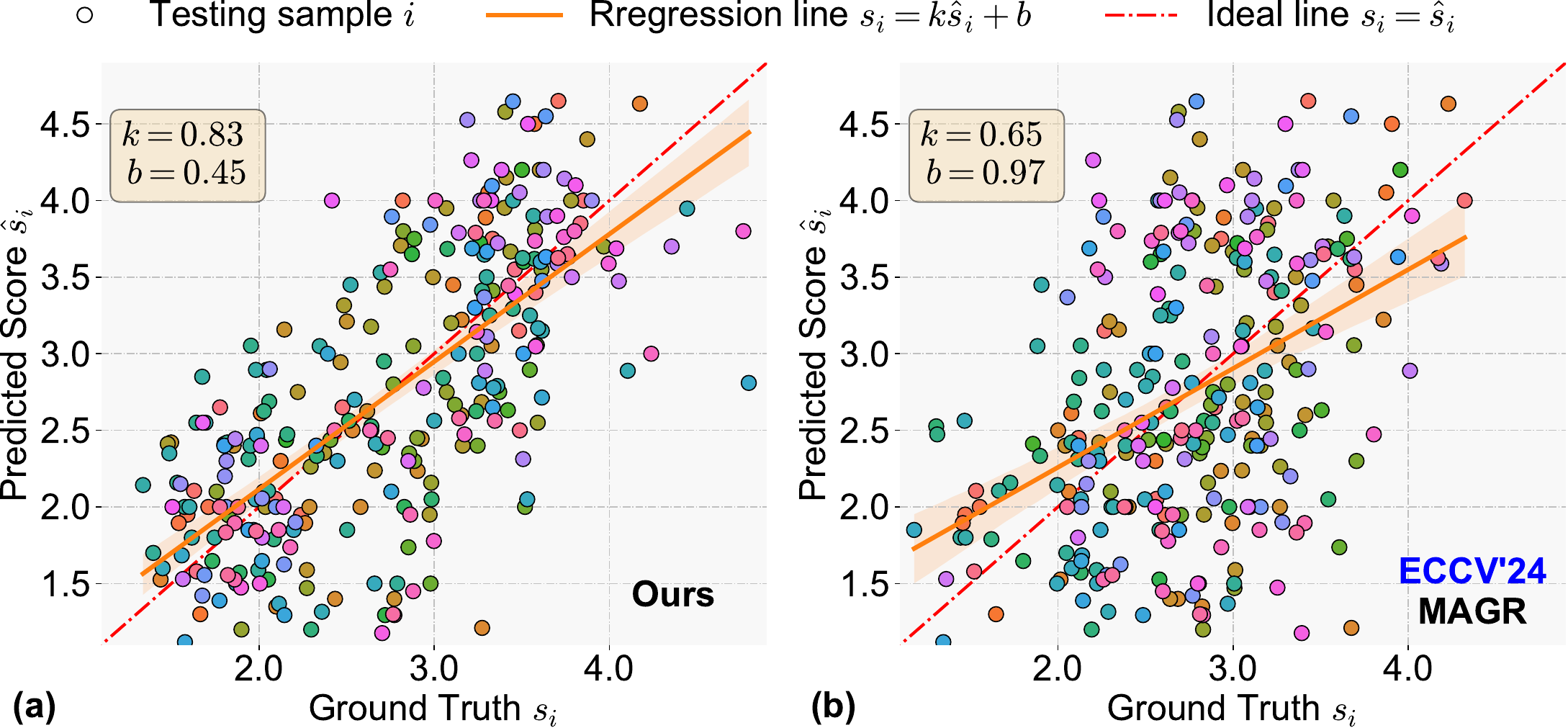}
    \caption{
Scatter plots with regression lines showing the alignment between predicted scores and ground truth: (a) Ours, (b) MAGR.
    }
    \label{fig:relation_cmp}
\phantomsubcaption\label{fig:relation_cmp-a}
\phantomsubcaption\label{fig:relation_cmp-b}
\end{minipage}~
\begin{minipage}{.34\textwidth}
    \centering
    \includegraphics[width=\linewidth]{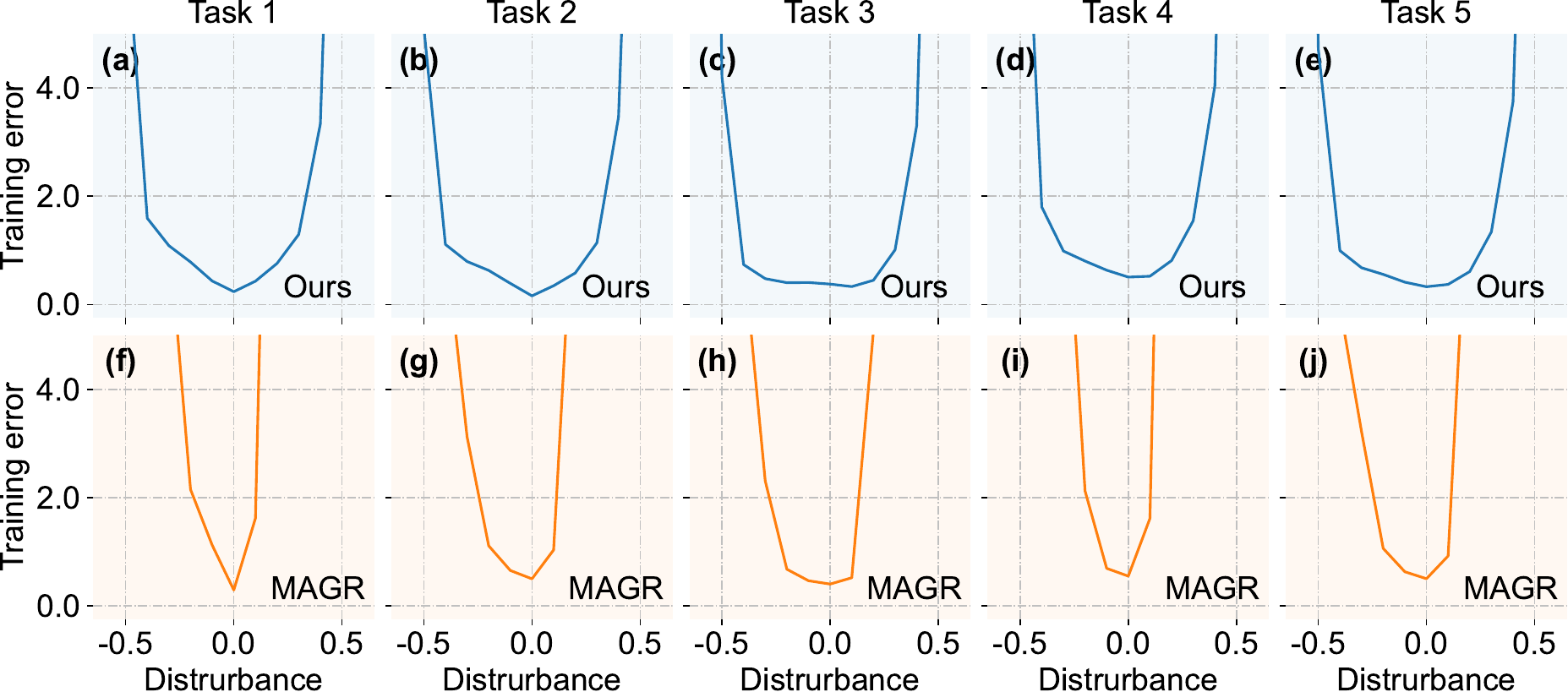}
    \caption{
    Curvature plots of the loss landscape around the obtained solution after continual learning of all tasks: (a-e) Ours, (f-j) MAGR. 
    }
    \label{fig:loss_landscape}
\phantomsubcaption\label{fig:loss_landscape-a}
\phantomsubcaption\label{fig:loss_landscape-b}
\phantomsubcaption\label{fig:loss_landscape-c}
\phantomsubcaption\label{fig:loss_landscape-d}
\phantomsubcaption\label{fig:loss_landscape-e}
\phantomsubcaption\label{fig:loss_landscape-f}
\phantomsubcaption\label{fig:loss_landscape-g}
\phantomsubcaption\label{fig:loss_landscape-h}
\phantomsubcaption\label{fig:loss_landscape-i}
\phantomsubcaption\label{fig:loss_landscape-j}
\end{minipage}
\end{figure*}

\subsubsection{Visualizations}
\paragraph{Case Study}
\cref{fig:case_study} illustrates various case comparisons of our ASAL model and \cite{wen2024perceptual}. For Sample \#0016 (see \cref{fig:case_study-a}) with a ground truth score of 1.56, our model’s prediction of 1.93 accurately identifies distortions such as abrupt structure changes and over-exposure, whereas \cite{wen2024perceptual} overestimates with a score of 3.93. In Sample \#1413 (see \cref{fig:case_study-b}), our model predicts 4.40, closely matching the ground truth of 4.50, while \cite{wen2024perceptual} slightly overestimates with a score of 4.72. In Sample \#1800 (see \cref{fig:case_study-c}), both models face challenges; our prediction of 2.10 and \cite{wen2024perceptual}’s 4.39 show significant errors for the ground truth score of 1.39, highlighting difficulties in detecting complex distortions. \cref{fig:case_study}  demonstrates the superiority of ASAL in handling various distortions.

\paragraph{Feature Distribution and Regression Alignment}
We provide the feature distribution and regression visualizations in \cref{fig:tsne_cmp,fig:relation_cmp} for a better understanding.
The t-SNE plots in \cref{fig:tsne_cmp} reveal that our model (see \cref{fig:tsne_cmp-a}) achieves a more distinct separation of testing samples compared to MAGR (see \cref{fig:tsne_cmp-b}). 
Here, we discretize the continuous score into five classes.
This enhanced separation indicates a more effective feature representation, allowing for better differentiation between varying scores. Additionally, the scatter plots with regression lines in \cref{fig:relation_cmp} demonstrate that the regression line $\hat{s}_i = 0.83s_i+0.48$ for our model (see \cref{fig:relation_cmp-a}) is closer to the ideal line $\hat{s}_i = s_i$ than that $\hat{s}_i = 0.65s_i+0.97$ of MAGR (see \cref{fig:relation_cmp-b}), suggesting superior alignment between predicted scores and ground truth. This closer alignment reflects improved prediction accuracy and the superiority of our model in capturing perceptual details for accurate assessment.


\paragraph{Flat Minima Test}
To validate the model generalization, we conducted a flat minima test in the loss landscape \cite{deng2021flattening}.
We analyze the change in training loss for each task by perturbing the weights in random directions. This process is repeated ten times, and the average loss across these disturbances is computed. 
The curvature plots in \cref{fig:loss_landscape} show the loss landscape around the obtained solution after continual learning of the final task for our model and MAGR \cite{zhou2024magr}. 
The loss landscape in all five tasks (see \cref{fig:loss_landscape-a,fig:loss_landscape-b,fig:loss_landscape-c,fig:loss_landscape-d,fig:loss_landscape-e}) indicates that our model has achieved a more robust solution, which is associated with better generalization and reduced risk of catastrophic forgetting. In contrast, MAGR shows sharper minima (see \cref{fig:loss_landscape-f,fig:loss_landscape-g,fig:loss_landscape-h,fig:loss_landscape-i,fig:loss_landscape-j}), suggesting less stability and a higher likelihood of poorer generalization and forgetting.
The improved generalization of our model is particularly beneficial in VR-VQA tasks where the content varies widely in VR environments.

\section{Conclusions and Discussions}
In this work, we propose ASAL as an innovative approach for the perceptual quality assessment of VR videos. ASAL improves VR-VQA models by incorporating novel strategies for enhancing generalization, effectively balancing correlation and precision for accurate assessment. Additionally, ASAL is extended with a continual learning framework, leveraging adaptive memory replay and feature adaptation to mitigate catastrophic forgetting and limited computational capacity in VR devices. Our ASAL method outperforms recent strong baselines in both the joint training and continual learning settings, showing significant improvements in the correlation between predicted and ground truth scores and reduced error rates. Overall, ASAL not only advances the state-of-the-art in perceptual quality assessment for 360-degree videos but also sets a robust benchmark for integrating continual learning in VR environments.
Further, our work paves the way for more adaptive, efficient, and user-centric VR applications in diverse domains.

For future work, several directions can be pursued to further advance the field. First, expanding the dataset to include more diverse and challenging VR video scenarios could improve model robustness and generalization. Second, exploring more advanced techniques for dynamic key frame selection and memory management could enhance the adaptability of the model to evolving content. Finally, integrating real-time feedback to continuously refine the model based on user interactions in VR represents a promising avenue for further research.

\acknowledgments{%
This work was supported in part by the National Natural Science Foundation of China under Grant 62272019.
}

\bibliographystyle{abbrv-doi-hyperref}

\bibliography{refs}

\end{document}